\documentclass{article}

\PassOptionsToPackage{numbers, sort, compress}{natbib}

\usepackage[final]{neurips_2025}

\usepackage[utf8]{inputenc} %
\usepackage[T1]{fontenc}    %
\usepackage{hyperref}       %
\usepackage{url}            %
\usepackage{booktabs}       %
\usepackage{amsfonts}       %
\usepackage{nicefrac}       %
\usepackage{microtype}      %
\usepackage{xcolor}         %
\usepackage{amsmath}
\usepackage{xspace}
\usepackage{graphicx}
\usepackage{comment}
\usepackage{amssymb}
\usepackage{algorithm}
\usepackage{enumitem}
\usepackage{algorithmic}
\usepackage[capitalize,noabbrev]{cleveref}
\usepackage{caption}
\usepackage{subcaption}
\usepackage[frozencache,cachedir=./_minted-main_arxiv]{minted}
\usepackage{adjustbox}

\definecolor{codebg}{rgb}{0.95, 0.95, 0.95}
\setminted{
    fontsize=\footnotesize,
    baselinestretch=1,
    linenos=true,
    breaklines=true,
    breakanywhere=true,
    frame=lines,
    framesep=2mm,
    bgcolor=codebg,
    tabsize=4,
    escapeinside=||,
}

\usepackage{tcolorbox}
\tcbuselibrary{listings,breakable} %
\newtcblisting{promptbox}[1][Prompt]{%
  listing only,
  listing options={
    basicstyle=\ttfamily\scriptsize,
    breaklines=true,
    prebreak=\mbox{\tiny\ensuremath\hookleftarrow},
  },
  title={#1},
  fonttitle=\bfseries,
  breakable,
}

\title{Once Upon an Input: Reasoning via Per-Instance Program Synthesis}

\newcommand*\samethanks[1][\value{footnote}]{\footnotemark[#1]}

\author{%
Adam Stein\thanks{These authors contributed equally to this work.}\qquad Neelay Velingker\samethanks\qquad Mayur Naik\qquad Eric Wong \\
\texttt{\{steinad, neelay, mhnaik, exwong\}@seas.upenn.edu} \\
University of Pennsylvania
}

\newcommand{\ourmethod}[1][\relax]{%
  \textsc{PIPS}%
  \ifx\relax#1\relax
  \else
    -#1%
  \fi
  \xspace
}

\begin{document}

\maketitle

\begin{abstract}
Large language models (LLMs) excel at zero-shot inference but continue to struggle with complex, multi-step reasoning. Recent methods that augment LLMs with intermediate reasoning steps such as Chain of Thought (CoT) and Program of Thought (PoT) improve performance but often produce undesirable solutions, especially in algorithmic domains.
We introduce Per-Instance Program Synthesis (\ourmethod), a method that generates and refines programs at the instance-level using structural feedback without relying on task-specific guidance or explicit test cases. To further improve performance, \ourmethod incorporates a confidence metric that dynamically chooses between direct inference and program synthesis on a per-instance basis.
Experiments across three frontier LLMs and 30 benchmarks including all tasks of Big Bench Extra Hard (BBEH), visual question answering tasks, relational reasoning tasks, and mathematical reasoning tasks show that \ourmethod  improves the absolute harmonic mean accuracy by up to 8.6\% and 9.4\% compared to PoT and CoT respectively,
and reduces undesirable program generations by 65.1\% on the algorithmic tasks compared to PoT with Gemini-2.0-Flash. \footnote{Code for experiments and a demo is open-sourced at \url{https://github.com/adaminsky/pips}.}
\end{abstract}
\section{Introduction}

Large-scale pretraining endows LLMs with the ability to recognize common concepts and perform many tasks in a zero-shot fashion
but they  still struggle with multi-step reasoning \citep{dziri2023faith, kambhampati2024position, zhou2024larger}. Recent advances in inference-time reasoning strategies such as Chain of Thought (CoT) and related work \citep{cot, openai2024learning, guo2025deepseek} have significantly improved LLMs' reasoning abilities.
However, they remain unreliable \citep{bbeh, chen2024not, sprague2024musr, stechly2024chain} and unfaithful, meaning the final answer is correct for the wrong reasons \citep{lanham2023measuring, chen2025reasoning, turpin2023language}.

Unlike LLM inference, program execution enforces precise, verifiable computation.
Combining LLMs with program execution offers a promising reasoning method: the LLM handles
perceptual inference, mapping raw input to structured form, while algorithmic reasoning is offloaded to an executable program \citep{kambhampati2024position, vieira, wang2024hypothesis}.
Existing work on \textit{neuro-symbolic learning} as well as methods such as Faithful Chain of Thought (FCoT) \citep{fcot} and Program Aided Language Models (PAL) \citep{pal} adopts this approach, however, they use a single fixed program per task which causes problems when task instances are varied \citep{stein2025road}.
On the other hand, methods such as Program of Thought (PoT) \citep{pot} aim to enable LLMs to generate these programs in a zero-shot manner on an \textit{instance-level}.

While flexible, these methods often produce undesirable programs, due to three challenges in instance-level program synthesis:
(1)~\textit{open domain}: determining for a given instance if program synthesis is preferable to direct inference (via CoT) remains an open question,
(2)~\textit{no task specifications}: there are no general specifications for how the correct program should behave to guide program search, and
(3)~\textit{unstructured input}: programs typically operate over structured input, but common reasoning problems are unstructured, requiring on-the-fly instance-level input understanding.

\begin{figure}[t]
    \centering
    \includegraphics[width=\linewidth]{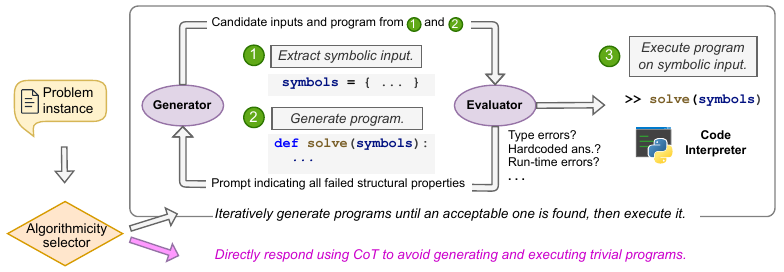}
    \vspace{-0.15in}
    \caption{Overview of Per-Instance Program Synthesis (\ourmethod). \ourmethod addresses the open-domain nature of reasoning problems by selecting between synthesis and CoT at the instance-level, avoiding unnecessarily generating programs for non-algorithmic problems. For algorithmic problems, \ourmethod addresses the lack of task specifications by iteratively synthesizing programs using feedback based on \textit{structural checks}. \ourmethod handles unstructured input via instance-specific symbolic extraction (step~1) before program synthesis (step~2). \cref{fig:synth-loop} shows an example where an undesirable program is rejected before producing an acceptable one which gives the correct answer upon execution (step~3).}
    \label{fig:overview}
    \vspace{-0.05in}
\end{figure}

In this paper, we propose Per-Instance Program Synthesis, or \ourmethod, to solve reasoning problems at the instance-level. \Cref{fig:overview}
illustrates how \ourmethod addresses the aforementioned challenges.
To address the open domain nature of such problems, \ourmethod introduces an instance-level confidence metric to decide if the LLM is better suited to solving the instance with program synthesis or direct inference.
It then iteratively generates and evaluates programs using feedback based on structural checks (e.g. non-triviality, syntax, type errors), specifically designed to avoid the collapse to trivial solutions and ensure well-formed computation.
To handle unstructured input, \ourmethod explicitly performs instance-specific symbolic extraction before synthesis, decoupling program search from perceptual inference.
Unlike other iterative refinement and debugging methods for code generation \citep{gupta2020synthesize, wang2024hypothesis, chen2024teaching}, \ourmethod does not require explicit test cases, examples, or other forms of task specifications.

Our experiments demonstrate that \ourmethod significantly reduces the amount of undesirable programs produced compared to baselines, and this results in improved accuracy. Notably, \ourmethod reduces undesirable programs by 65.1\% for algorithmic benchmarks in the Big Bench Extra Hard (BBEH) suite~\citep{bbeh} as well as 7 additional tasks including visual question answering and mathematical reasoning, resulting in an 8.6\% absolute improvement in harmonic mean accuracy over PoT.
We also show that our confidence metric allows us to correctly switch between CoT and program synthesis for 65\% of cases, resulting in \ourmethod matching CoT accuracy for majority non-algorithmic tasks and even improving performance on majority algorithmic tasks.

Our contributions are as follows:
\begin{itemize}[leftmargin=*,itemsep=0pt,topsep=0pt]
    \item Per-Instance Program Synthesis (\ourmethod): An iterative program synthesis method guided by instance-specific feedback on program structure properties, improving reasoning by addressing the challenges of per-instance program synthesis methods.
    \item Synthesis Confidence Metric: 
    We study the tradeoff between program synthesis and CoT for answering reasoning problems, and we propose a synthesis confidence metric which predicts, prior to generation, which approach is more likely to yield a correct solution for a given model.
    \item State-of-the-art accuracy: \ourmethod improves code utilization by 65.1\% on algorithmic questions leading to an 8.6\% improvement in harmonic mean accuracy over PoT across 30 frontier tasks.
\end{itemize}

\begin{figure}[t]
    \centering
    \includegraphics[width=\linewidth]{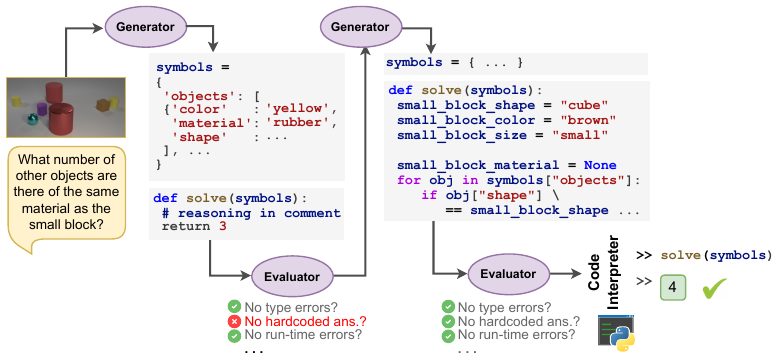}
    \caption{Example illustrating two iterations of the synthesis loop in \ourmethod.
    }
    \label{fig:synth-loop}
\end{figure}
\section{The Challenges of Per-Instance Program Synthesis}

A reasoning problem is a pair $(x, y)\in \mathcal{X}\times\mathcal{Y}$ where $x$ is a raw query consisting of unstructured data such as natural language text or an image, and $y$ is the answer which we assume is represented symbolically (something that could be output by a program).
Existing methods for generating instance-level programs for solving reasoning problems, such as PoT, produce a program $P:\varnothing\to\mathcal{Y}$ to compute the correct answer. These programs take no input, similar to a main function, but can be highly general since they are often generated in a Turing-complete language such as Python.

This problem-solving strategy naturally introduces three  challenges.
In short, algorithmic problems are well-suited to program synthesis but non-algorithmic problems are inherently ill-suited;
producing a program cannot be guided by traditional specifications;
and interfacing the unstructured input from the problem with the program requires on-the-fly instance-level perceptual understanding.

\subsection{Non-Algorithmic Problems}
\label{sec:open-domain}

\begin{figure}[ht]
    \centering

    \begin{subfigure}[t]{0.46\textwidth}
        \centering
        \includegraphics{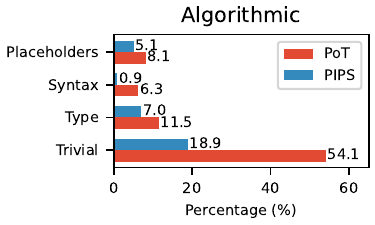}
        \caption{Fraction of output code with issues:
  \textit{trivial} (hard-coded answer), \textit{type} (wrong return type), \textit{syntax} (parser error), and \textit{placeholders} (incomplete code).}
        \label{fig:code-issues}
    \end{subfigure}
    \hfill
    \begin{subfigure}[t]{0.52\textwidth}
        \centering
        \includegraphics{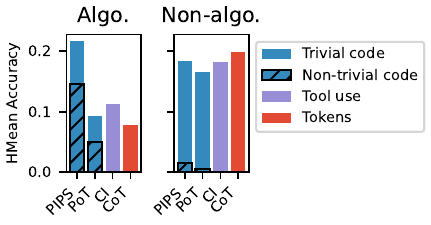}
        \caption{Per-instance synthesis baseline performance on mostly algorithmic vs. non-algorithmic problems. Shading indicates fraction of correct solutions with non-trivial code.}
        \label{fig:algorithmic-trivial}
    \end{subfigure}

    \caption{Failures in existing approaches to per-instance program synthesis with Gemini-2.0-Flash.}
    \label{fig:pot-issues}
\end{figure}

The per-instance synthesis setting poses a tradeoff between program synthesis and direct inference reliability. In some cases, it is more reliable for an LLM to solve a problem through token-based reasoning (like CoT) than through program synthesis.
Examples include tasks like emotion understanding and summarization, which are not traditionally considered ``algorithmic''.
We find that when existing per-instance synthesis methods are applied to these non-algorithmic settings, they will often output trivial programs that unnecessarily invoke the Python interpreter.

To study the \textit{algorithmicity} of the problem instances across a variety of datasets, we design an LLM-based classifier to determine if an instance is algorithmic, meaning it could be solved by executing a non-trivial program.
Using this classifier, we split our suite of datasets (which includes all 23 tasks of BBEH as well as 7 additional tasks) into {\em algorithmic} ones which have a majority of algorithmic instances, and {\em non-algorithmic} ones which have a majority of non-algorithmic instances.
Details of our classifier and the full task split is provided in \cref{app:dataset-split}. Overall, we find that not all tasks are purely algorithmic or non-algorithmic, meaning that algorithmicity is best determined on the instance-level. In addition, 
as Figure~\ref{fig:algorithmic-trivial} shows, code execution via PoT prompting significantly improves performance on the algorithmic tasks, while having a slightly negative impact for the non-algorithmic tasks. 
We follow the recommendation from \citet{bbeh} to evaluate harmonic mean accuracy since it is a challenging metric that requires improvements on the hardest tasks to improve aggregate performance. For these non-algorithmic problems, PoT rarely outputs well-formed code.
Instead of producing superficial code for non-algorithmic problems, we could skip program synthesis entirely with no harm to performance or interpretability.

\subsection{Program Search without Specifications}

\begin{figure}[t]
    \centering
    \begin{subfigure}[b]{0.49\linewidth}
        \centering
        \includegraphics[width=\linewidth]{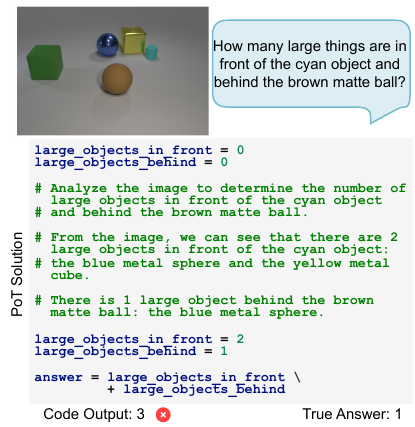}
        \caption{Trivial program.}
        \label{fig:trivial-program}
    \end{subfigure}
    \hfill
    \begin{subfigure}[b]{0.49\linewidth}
        \centering
        \includegraphics[width=\linewidth]{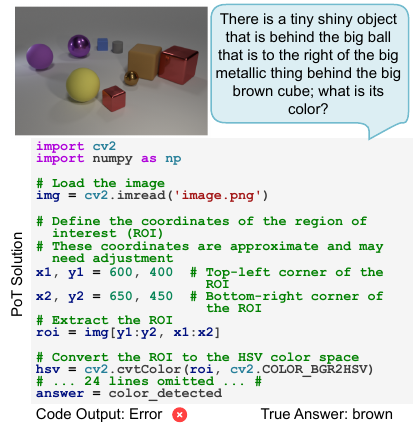}
        \caption{Program operating over raw image.}
        \label{fig:opencv}
    \end{subfigure}
    \caption{Two programs generated with PoT illustrating program synthesis failures. Part (a) shows a trivial program where two variables are initialized to zero, but then several steps of reasoning are performed in comments, leading to their values being hard-coded rather than computed with code. Part (b) shows an input-free program to process the input image itself which would be better done using the LLM's perceptual inference. Both programs result in the wrong answer. The corresponding programs produced by \ourmethod yield the correct answer in both cases and are shown in \cref{app:our-programs}.}
    \label{fig:synthesis-failures}
\end{figure}

Existing instance-level code generation methods often use the first generated program \citep{pot}, or use minimal forms of program search due to the lack of any explicit specifications for how the correct program should behave.
We find that this frequently leads to ``trivial'' programs which have an explicit return value hard-coded. \cref{fig:code-issues} shows that over 50\% of PoT's outputs on algorithmic tasks fall into this category.
The lack of behavioral specifications leaves the LLM the option of solving the task through direct inference with the final answer wrapped in a program. \Cref{fig:trivial-program} shows a trivial program produced by PoT. Additionally, 6.3\% of programs have syntax errors and 11.5\% return the wrong type which is why it is undesirable to always settle with the first generated program. Per-instance synthesis approaches suffer from these problems since they lack the necessary task specifications to perform traditional program search. Our full evaluation criteria including well-formed programs, type errors, and syntax errors is in \cref{app:eval-criteria}.

\subsection{Interfacing Programs with Unstructured Data}
\label{sec:unstructured}

The use of programs is best when their input is a well-defined symbolic structure as opposed to raw data (e.g. a paragraph of text or an image).
Since existing methods generate input-free programs, they must either hard-code the necessary structured input into the program (using the LLM's perceptual understanding) or use the code to process the unstructured input.
\Cref{fig:opencv} shows an example of a program produced by PoT on the CLEVR task \citep{clevr} wherein the program attempts to parse objects from an image using code instead of leveraging the strong perceptual understanding abilities of the LLM itself.
Executing the program leads to an error caused by referencing a non-existent file.

We find that 12.7\% of well-formed PoT code solutions to the CLEVR and Leaf multimodal datasets use the OpenCV or Pillow libraries, representing a fundamentally brittle approach to input understanding.

\section{Per-Instance Program Synthesis (\ourmethod)}
\label{sec:method}
This section addresses the above three challenges with an approach to synthesize programs on a per-instance basis without task specifications.
We use the general problem-solving structure of $y = P(c(x))$ where $c: \mathcal{X}\rightarrow\mathcal{R}$ is a function mapping from raw input space to permissible program inputs and $P: \mathcal{R}\rightarrow\mathcal{Y}$ is a program in a Turing-complete programming language. The next sections describe how we address the previous challenges with this problem-solving framework.

\subsection{Selective Program Synthesis}
While program synthesis allows for more faithful and sophisticated reasoning, there may still be specific problem instances where CoT should be used.
The decision between synthesis and CoT for solving an instance depends on both the algorithmic nature of the problem and the model’s capabilities in the two respective solving approaches.
Formally, given a reasoning problem $(x, y) \in \mathcal{X} \times \mathcal{Y}$, we must choose between two strategies: (1) direct CoT reasoning via $\hat{y} = M_{\text{cot}}(x)$, where $M_{\text{cot}}$ denotes chain of thought inference, or (2) program synthesis via $\hat{y} = P(c(x))$, where $P$ is a synthesized program and $c(x)$ is a symbolic abstraction of the input $x$.

Since the decision depends on the model’s own problem-solving abilities, we elicit it from the model before it begins reasoning. Our self-prompting method uses ten criteria to choose between CoT and program synthesis. Each criterion results in a confidence score from the model, forming a vector $S(x) = (p_1(x), \dots, p_{10}(x)) \in [0,1]^{10}$. These criteria assess factors like ease of formalization, expected execution success, and robustness of logic. For reasoning models, we include ten additional criteria relating to their reasoning abilities.
Motivated by prior work demonstrating that LLMs can accurately estimate the likelihood that their answer to a question is correct \citep{kadavath2022language}, we hypothesize that these criteria, which are agnostic to the downstream task, can be strong predictors for the LLMs success in per-instance synthesis.
Full details of the criteria are in \cref{app:switch-criteria}.

Given the model's own assessment of its abilities through these probing questions, the final switch decision can either be derived in a fully zero-shot manner, or a held-out calibration set can be used to derive the decision from a learned logistic classifier, enabling the switch to leverage problem solving experience for higher accuracy.

\subsection{Program Synthesis without Task Specifications}
\label{sec:synthesis}

\begin{algorithm}[t]
\caption{\ourmethod: Synthesis Loop}
\label{alg:ourmethod}
\begin{algorithmic}[1]
\small
\REQUIRE Input $x \in \mathcal{X}$, symbolic extractor $c$, maximum iterations $k$
\ENSURE Program $P^*$ such that $P^*(c(x)) \approx y$
\STATE Initialize $i \leftarrow 0$
\STATE Extract symbols: $r_0 \leftarrow c(x)$
\STATE Generate initial program: $P_0 \leftarrow \texttt{LLM}(x, r_0)$
\FOR{$i=1$ to $k$}
    \STATE Evaluate program: $F_i \leftarrow E(P_i, r_i, x)$
    \IF{$F_i = \texttt{Pass}$}
        \STATE \textbf{return} $P_i$
    \ENDIF
    \STATE Update symbols if they have associated errors: $r_{i+1} \leftarrow c(x; \{r_j\}_0^{i}, \{P_j\}_0^i, \{F_j\}_0^i)$
    \STATE Generate revised program: $P_{i+1} \leftarrow \texttt{LLM}(x; \{r_j\}_0^{i+1}, \{P_j\}_0^i, \{F_j\}_0^i)$
\ENDFOR
\STATE \textbf{return} $P_k$ \COMMENT{Fallback if none pass}
\end{algorithmic}
\end{algorithm}
 
Traditional program synthesis searches the space of programs guided by a task specification (either input-output examples or a logical specification) which determine which programs are acceptable and which are not. To perform program synthesis with just a single input, we leverage auxiliary forms of specifications based on generally undesirable failure modes of code generation. Formally, given an input $x\in\mathcal{X}$ which may consist of a request such as ``How is Bob related to Alice'' as well as image inputs like an image of a family tree, we want to find a program $P$ such that $P(c(x))=y$ where $y$ is the true answer. Our goal is to search for a $P$ to optimize the following:
\begin{align*}
    \min_P M(P \mid x, c(x)) + \lambda E(P, c(x); x)
\end{align*}
where $M$ is an LLM and $E$ is a program evaluator which checks various properties of code based on static and dynamic analysis. To design the program evaluator $E$, we study the common failure modes of LLM code generation in this setup without a feedback loop.

\ourmethod{} is an approach to solve for $P$, described in \cref{alg:ourmethod} and illustrated in \cref{fig:overview}.
We design $E$, consisting of an LLM and a program interpreter, to flag any program $P_i$ matching any of the aforementioned patterns. 
If an issue is detected, $E$ produces structural feedback to aid in fixing the problem.
Issues with the structured input $r_i$ result in a revised program input $r_{i+1}$ conditioned on the history of programs and inputs. If any of the evaluator feedback pertains to the synthesized program, then the generator produces
a revised program $P_{i+1}$ conditioned on the feedback past programs. 
This iterative refinement continues for at most $k$ steps, or until $E(P_i, r_i, x)$ detects no issues.
We note that at the initial step $k=0$, $M$ is prompted to produce a program that does not contain these patterns.

\subsection{Converting Raw Data to Symbolic Program Input}

To address the issue of interfacing discrete programs with unstructured data, we explicitly abstract the unstructured data as structured program input before program generation. The input abstraction is done with the mapping $c$, and we use an LLM for this task due to their ability for understanding general unstructured data.
Concretely, the LLM processes the input to identify salient entities, their attributes, and the relationships between them that are pertinent to solving the instance. This requires the LLM to infer an ad hoc schema for the JSON structure, tailored to the specific semantics of the input, rather than have this schema be predetermined.
While allowing for an LLM to decide how to abstract the input into a structured form is highly general, this can lead to mistakes or missed information from the input in this step. The synthesis loop described above iteratively fixes such issues in the data abstraction in coordination with program generation.
\section{Experiments}
\label{sec:experiments}
This section studies whether \ourmethod addresses the challenges of per-instance program synthesis. \textbf{RQ1} studies the empirical performance of \ourmethod, and \textbf{RQ2}, \textbf{RQ3}, and \textbf{RQ4} correspond to each respective challenge laid out in \cref{sec:method}.

\begin{figure}[t]
    \centering
    \includegraphics{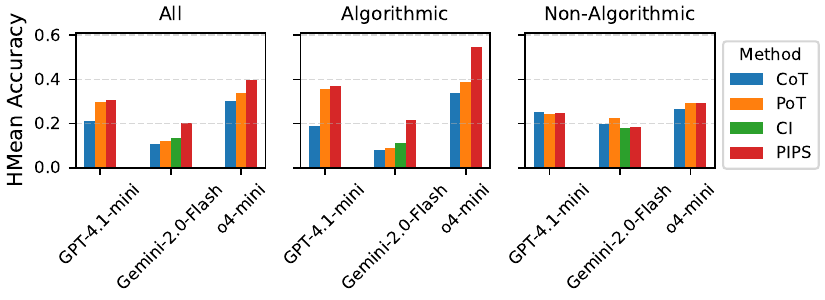}
    \caption{Harmonic mean accuracy over all 30 datasets (left), on the 17 majority algorithmic tasks (middle), and on the 10 majority non-algorithmic (right) for \ourmethod and baselines using three state-of-the-art models.
    The breakdown per task per model is shown in \cref{tab:gemini_20_flash_details}, \cref{tab:o4_mini_2025_04_16_details}, and \cref{tab:gpt_41_mini_2025_04_14_details}.}
    \label{fig:all-results}
\end{figure}

\subsection{Setup}
\textbf{Datasets} %
We evaluate our approach using 23 tasks sourced from the Big Bench Extra Hard (BBEH) benchmark \citep{bbeh}. These tasks span topics such as geometric understanding, deductive logical reasoning, and commonsense understanding. Furthermore, we extend this study to the visual reasoning tasks CLEVR \citep{clevr} and Leaf \citep{solkobreslin2024dataefficient}, the relational reasoning task CLUTTR \citep{clutrr}, and four mathematical reasoning tasks of OmniMath \citep{omnimath}. For all datasets, we reserve a random sample of 20\% of the data for calibration of our confidence switch, and we evaluate on the remaining 80\% of the data. We also evaluate the generalizability of a trained confidence switch in a leave-one-dataset out evaluation scheme as well as show that a fully zero-shot confidence switch is also highly effective in \cref{app:switch-ablations}.

\textbf{Models} 
To evaluate our setup across a variety of different frontier LLMs, we use Gemini-2.0-Flash \citep{team2023gemini}, GPT-4.1-mini \citep{openai2025gpt41mini}, and o4-mini \citep{openai2025o3o4mini}. Gemini-2.0-Flash is a multimodal LLM, GPT-4.1-mini is a lightweight general-purpose LLM, and o4-mini is a multimodal LLM that was trained for ``reasoning'' with a long CoT before its final response.

\textbf{Baselines} We evaluate \ourmethod{} against Program of Thought (PoT) \citep{cot}, Chain of Thought \citep{cot}, and a code interpreter tool-use agent. PoT involves prompting an LLM to generate Python code to solve the problem and then executing the code to get a final answer.
Gemini Code Interpreter (CI) \citep{team2023gemini} is an API tool that allows the model to synthesize and execute code in an \textit{agentic} manner before producing its final response.

We evaluate another agentic baseline called PoT-retries which performs PoT, but regenerates the code until the code executes without any errors, as well as CodeAct \citep{wang2024executable}, and Buffer-of-Thoughts \citep{yang2024buffer} for Gemini-2.0-Flash in \cref{app:more-baselines}.

\subsection{RQ1: Does \ourmethod outperform baselines across datasets?}

\begin{figure}[t]
    \centering
    \begin{subfigure}[t]{0.49\textwidth}
        \centering
        \includegraphics{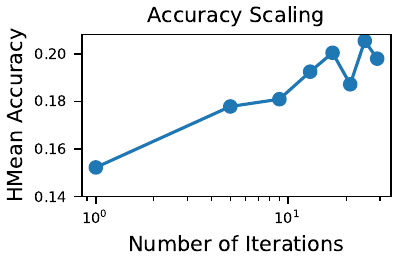}
        \caption{Scaling of harmonic mean accuracy across all datasets with more synthesis iterations.}
        \label{fig:acc-scaling}
    \end{subfigure}
    \hfill
    \begin{subfigure}[t]{0.49\textwidth}
        \centering
        \includegraphics{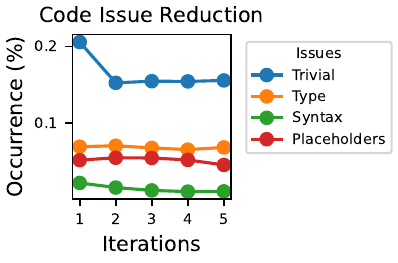}
        \caption{Decrease in code issues with more synthesis iterations.}
        \label{fig:code-scaling}
    \end{subfigure}
    \caption{Accuracy and code quality scaling with more iterations of \ourmethod with Gemini-2.0-Flash.}
    \label{fig:combined}
\end{figure}

We compare \ourmethod to baselines in terms of overall performance.
Harmonic mean aggregated results of \ourmethod compared to baselines over all 30 datasets as well on just the algorithmic and non-algorithmic tasks are shown in \cref{fig:all-results}.
Overall, we see an absolute improvement of 8.6\% in harmonic mean accuracy over PoT, with up to a 23.7\% improvement in absolute accuracy over PoT (on BBEH Boolean Expressions) for Gemini-2.0-Flash, a 0.8\% absolute improvement over PoT for GPT-4.1-mini, and a 5.7\% absolute improvement over PoT for o4-mini.
From the middle plot, we can see that \ourmethod provides significant improvements over the baselines (up to 15.9\% in absolute harmonic accuracy for o4-mini) on the majority algorithmic problems, while not degrading in accuracy for the non-algorithmic tasks.
Full results are included in \cref{app:full-results}.

We further study the performance of \ourmethod as the number of feedback iterations $k$.
Even at $k=0$, meaning the evaluator is not used, \ourmethod{} outperforms PoT by a harmonic mean difference of 5.6\% and CI by 3.7\%.
This gap widens as we scale $k$, as shown in \cref{fig:acc-scaling}.
Iteration successfully leads to more well-formed programs which subsequently improves correctness.

We report the average cost of \ourmethod and baselines in \cref{app:cost} and show that \ourmethod additionally achieves lower cost than other iterative approaches.

\begin{table}[h]
\centering
\small
\caption{Ablation study for BBEH tasks.}
\label{tab:ablation_results}
\begin{tabular}{lcc}
\toprule
Method & HMean Accuracy (\%) \\
\midrule
\ourmethod & 20.8 \\
\ourmethod (no switch) & 18.3 \\
\ourmethod[0] (no switch) & 12.9 \\
\ourmethod[0] (no switch, no symbols) & 4.3 \\
\bottomrule
\end{tabular}
\end{table}

\subsection{RQ2: How effective is switching between synthesis and CoT on the instance-level?}

In this RQ, we investigate whether our switch can effectively decide between synthesis and CoT before committing to either option.
As described in \cref{sec:open-domain}, this is important since non-algorithmic problems almost always result in trivial programs which are equivalent to CoT, but incur an unnecessary call to the Python interpreter. Note that non-algorithmic instances occur even in majority algorithmic tasks. We show in \cref{app:pips-nonalgo} that encouraging non-trivial code for non-algorithmic instances with our iterative search process can reduce performance.

To evaluate our switch, we focus on cases where it affects the outcome—i.e., when either \ourmethod or CoT is correct, but not both. These comprise 24.8\% of all samples across 30 benchmarks. For Gemini-2.0-Flash, the switch selects the correct method 65.3\% of the time, yielding a 2.2\% absolute gain in harmonic mean accuracy (\cref{tab:ablation_results}). Example switch decisions are shown in \cref{app:switch-decisions}.

\begin{figure}
    \centering
    \includegraphics{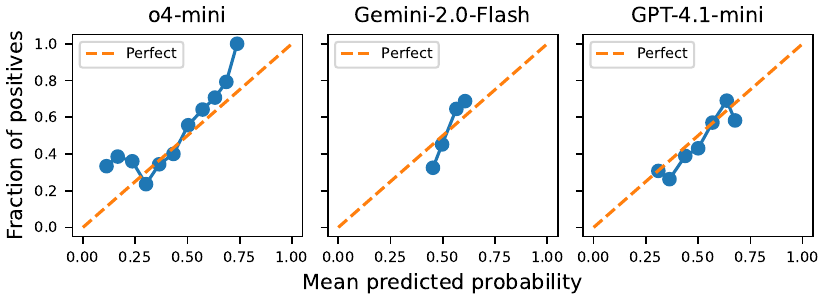}
    \caption{Calibration curve of our selection method between program synthesis and CoT. We only consider questions where the choice between synthesis and CoT determines answer correctness, and a positive instance is one correctly solved by code. Therefore, a score of 0.8 should mean an 80\% chance of solving correctly with code and a 20\% chance of solving correctly with CoT.}
    \label{fig:calibration}
\end{figure}

Furthermore, we investigate the level of calibration of our switching method.
As illustrated in Figure \ref{fig:calibration}, the switch is indeed well calibrated.
Notable sources of deviation occur at the extremes.
We further study the marginal contributions of each $p_i$ with respect to $S$ in \cref{app:switch-questions}.
Importantly, these results demonstrate the usefulness of an LLM's intrinsic understanding of its own problem solving abilities for choosing when program synthesis is appropriate on an instance-level.

\subsection{RQ3: Does \ourmethod improve code quality and correctness?}

\begin{table}[h]
\centering
\small
\caption{Performance boost from PIPS fixing PoT's code issues on BBEH algorithmic tasks. Boost reflects accuracy gain on samples where PIPS corrected PoT errors.}
\label{tab:pips_fixes_issues}
\begin{tabular}{l r r}
\toprule
Issue Fixed by \ourmethod & Performance Boost ($\Delta$ Acc \%) & Samples Fixed \\
\midrule
Syntax Errors & 20.0 & 200 \\
Wrong Return Type & 16.8 & 297 \\
Placeholders & 7.2 & 194 \\
Hardcoded Answers & 5.2 & 1138 \\
\bottomrule
\end{tabular}
\end{table}

In this RQ, we seek to study how \ourmethod improves code quality.
First, we analyze the performance results to ensure that \ourmethod produces programs that are meaningful.
To verify alignment, we filter \ourmethod and all baselines' code outputs by those that are well-formed according to our evaluator criteria described in \cref{sec:synthesis}.
As seen previously in \cref{fig:algorithmic-trivial}, \ourmethod produces significantly more well-formed programs than PoT when solving algorithmic problems, 
and the discrepancy in absolute percentage points can go up to 53.7\% as seen on the Temporal Sequence task. See \cref{app:full-results} for the percent of well-formed programs produced by \ourmethod and PoT for each of the 30 datasets.
Table \ref{tab:pips_fixes_issues} studies the marginal impact of each type of fix on BBEH tasks.

Next, we focus mainly on questions that are considered algorithmic, as decided by the classifier described in \cref{sec:open-domain}.
Focusing only on algorithmic samples across all benchmarks, Figure \ref{fig:code-issues} demonstrates that \ourmethod can significantly reduce the issues exhibited in PoT. 
For example, the number of trivial programs is reduced by as much as 75.6\%. 
Type and syntax issues are reduced by 49.2\% and 86.8\%, respectively. 
Lastly, the number of programs with placeholders are reduced by 36.3\%.
We study this more closely as iterations scale in Figure \ref{fig:code-scaling}, where we show as $k$ increases from 1 to 5, the percent occurrence of these undesirable properties decreases.

\subsection{RQ4: Does \ourmethod reduce ineffective handling of structured data?}

Unlike existing per-instance code generation methods, \ourmethod produces programs that take an explicit structured input. For instance, for image input, \ourmethod may extract relevant objects from the image and pass these objects to the program as a list. To determine if this explicit separation of the data and logic of a program reduces issues relating the incorrect handling of structured data, like that shown in \cref{fig:opencv}, we perform a comparison of the code produced by \ourmethod with PoT.
While 12.7\% of PoT well-formed code to the multimodal benchmarks (CLEVR and Leaf) used the OpenCV or Pillow libraries which are for image processing, our method never tries to manually process images.

We also perform an ablation as shown in the bottom two rows of \cref{tab:ablation_results}, where we see that the use of explicit function inputs in \ourmethod leads to a 4\% harmonic mean improvement on BBEH. Without explicit inputs, \ourmethod[0] (no symbols), produces a single input-free function, but first performing structured input extraction before code generation has a significant performance improvement.
\section{Related Work}

\textbf{Reasoning with Code Generation.}
LLMs have been used to generate structured symbolic representations—such as semantic parses \citep{fcot,vieira,hao2025planning} or domain-specific programs in PDDL, SMT, or Datalog—to enable external reasoning. These approaches rely on hand-crafted prompts and fixed DSLs that limit generality and expressiveness. Others prompt LLMs to produce executable code in general-purpose languages to solve problems directly \citep{pal,pan2023logiclm,pot}, enabling stronger abstraction and reuse. However, such methods typically rely on few-shot prompts or example-based verification (as in Programming-by-Example) \citep{li2024is,wang2024hypothesis,zelikman2023parsel}, limiting their applicability to tasks with clear specs or test cases. In contrast, \ourmethod performs instance-level program synthesis without requiring DSLs, specs, or handcrafted templates, and uses structural feedback to iteratively refine programs.

Approaches which prompt an LLM to produce code to solve a problem have been used in several domains beyond math and text-based reasoning questions. ViperGPT and followup work tackle visual question answering problems \citep{suris2023vipergpt, proptest}, Voyager applies to game playing \citep{wang2024voyager}, and Code as Policies focuses on the application of robot control \citep{liang2022code}. Recently, general systems such as CodeAct \citep{wang2024executable} and OpenCodeInterpreter \citep{opencodeinterpreter} have been proposed for solving problems via code generation similar to the previously mentioned solutions for each task.

\textbf{Test-Time Optimization for Reasoning.}
Prompting strategies like Chain of Thought \citep{cot,kojima2022large} and Tree of Thought \citep{yao2023tree} enhance LLM reasoning by decomposing problems or exploring multiple inference paths. Methods such as Hypothesis Search \citep{wang2024hypothesis} and Self-Discover \citep{zhou2024self} further improve performance by searching over program hypotheses or reasoning formats. Recent work also explores LLMs as code evaluators \citep{lightman2024lets,zaremba2014learning}, but often requires human supervision or task-specific tuning. \ourmethod differs by selecting between direct inference and code execution using a learned confidence signal, achieving robust and adaptable test-time reasoning with minimal assumptions.
 
\section{Limitations and Conclusion}
\label{sec:limitations}

In this paper, we focus on simple structural code properties since they occur often in generated code. Further work is needed to determine if there are more undesirable patterns in LLM-generated code. In addition, \ourmethod does not optimally handle problems which are best solved partly with CoT and partly with program synthesis. Future work can tackle methods for problem decomposition and composing program synthesis with other forms of reasoning. Finally, while \ourmethod offers interpretable reasoning when using code, the conversion of the input to symbolic form still lacks faithfulness guarantees.

We introduced Per-Instance Program Synthesis (\ourmethod), a method that dynamically synthesizes reasoning programs by leveraging general structural feedback. By focusing synthesis on the instance-level, rather than the task-level, \ourmethod significantly outperforms prior code-based reasoning approaches such as PoT as well as purely textual reasoning via CoT, and produces much less \textit{trivial} code than prior work. The efficacy of \ourmethod for solving the most challenging reasoning problems through program synthesis underscores the promise of synthesis as a powerful means of enabling complex reasoning, in addition to the current paradigm of CoT-driven reasoning.
 
\begin{ack}
We thank Mayank Keoliya for his feedback and help with additional baselines during the rebuttal period.

This research was supported by the ARPA-H program on Safe and Explainable AI under the award D24AC00253-00, an NSF Graduate Research Fellowship, a Google Research Award, and a gift from AWS AI to ASSET (Penn Engineering Center on Trustworthy AI).
\end{ack}

\bibliography{refs}
\bibliographystyle{unsrtnat}

\appendix

\renewcommand{\thefigure}{\thesection.\arabic{figure}}
\renewcommand{\thetable}{\thesection.\arabic{table}}
\setcounter{figure}{0}
\setcounter{table}{0}

\newpage

\section*{NeurIPS Paper Checklist}

\begin{enumerate}

\item {\bf Claims}
    \item[] Question: Do the main claims made in the abstract and introduction accurately reflect the paper's contributions and scope?
    \item[] Answer: \answerYes{} %
    \item[] Justification: The paper's claims are supported in Sections 2 and 3 with experiments.
    \item[] Guidelines:
    \begin{itemize}
        \item The answer NA means that the abstract and introduction do not include the claims made in the paper.
        \item The abstract and/or introduction should clearly state the claims made, including the contributions made in the paper and important assumptions and limitations. A No or NA answer to this question will not be perceived well by the reviewers. 
        \item The claims made should match theoretical and experimental results, and reflect how much the results can be expected to generalize to other settings. 
        \item It is fine to include aspirational goals as motivation as long as it is clear that these goals are not attained by the paper. 
    \end{itemize}

\item {\bf Limitations}
    \item[] Question: Does the paper discuss the limitations of the work performed by the authors?
    \item[] Answer: \answerYes{} %
    \item[] Justification: Limitations are addressed in \cref{sec:limitations}.
    \item[] Guidelines:
    \begin{itemize}
        \item The answer NA means that the paper has no limitation while the answer No means that the paper has limitations, but those are not discussed in the paper. 
        \item The authors are encouraged to create a separate "Limitations" section in their paper.
        \item The paper should point out any strong assumptions and how robust the results are to violations of these assumptions (e.g., independence assumptions, noiseless settings, model well-specification, asymptotic approximations only holding locally). The authors should reflect on how these assumptions might be violated in practice and what the implications would be.
        \item The authors should reflect on the scope of the claims made, e.g., if the approach was only tested on a few datasets or with a few runs. In general, empirical results often depend on implicit assumptions, which should be articulated.
        \item The authors should reflect on the factors that influence the performance of the approach. For example, a facial recognition algorithm may perform poorly when image resolution is low or images are taken in low lighting. Or a speech-to-text system might not be used reliably to provide closed captions for online lectures because it fails to handle technical jargon.
        \item The authors should discuss the computational efficiency of the proposed algorithms and how they scale with dataset size.
        \item If applicable, the authors should discuss possible limitations of their approach to address problems of privacy and fairness.
        \item While the authors might fear that complete honesty about limitations might be used by reviewers as grounds for rejection, a worse outcome might be that reviewers discover limitations that aren't acknowledged in the paper. The authors should use their best judgment and recognize that individual actions in favor of transparency play an important role in developing norms that preserve the integrity of the community. Reviewers will be specifically instructed to not penalize honesty concerning limitations.
    \end{itemize}

\item {\bf Theory assumptions and proofs}
    \item[] Question: For each theoretical result, does the paper provide the full set of assumptions and a complete (and correct) proof?
    \item[] Answer: \answerNA{} %
    \item[] Justification: There are no theoretical results presented in the paper.
    \item[] Guidelines:
    \begin{itemize}
        \item The answer NA means that the paper does not include theoretical results. 
        \item All the theorems, formulas, and proofs in the paper should be numbered and cross-referenced.
        \item All assumptions should be clearly stated or referenced in the statement of any theorems.
        \item The proofs can either appear in the main paper or the supplemental material, but if they appear in the supplemental material, the authors are encouraged to provide a short proof sketch to provide intuition. 
        \item Inversely, any informal proof provided in the core of the paper should be complemented by formal proofs provided in appendix or supplemental material.
        \item Theorems and Lemmas that the proof relies upon should be properly referenced. 
    \end{itemize}

    \item {\bf Experimental result reproducibility}
    \item[] Question: Does the paper fully disclose all the information needed to reproduce the main experimental results of the paper to the extent that it affects the main claims and/or conclusions of the paper (regardless of whether the code and data are provided or not)?
    \item[] Answer: \answerYes{} %
    \item[] Justification: We provide experimental details including the datasets and models used in \cref{sec:experiments} and we provide the prompts used in \cref{app:prompts}.
    \item[] Guidelines:
    \begin{itemize}
        \item The answer NA means that the paper does not include experiments.
        \item If the paper includes experiments, a No answer to this question will not be perceived well by the reviewers: Making the paper reproducible is important, regardless of whether the code and data are provided or not.
        \item If the contribution is a dataset and/or model, the authors should describe the steps taken to make their results reproducible or verifiable. 
        \item Depending on the contribution, reproducibility can be accomplished in various ways. For example, if the contribution is a novel architecture, describing the architecture fully might suffice, or if the contribution is a specific model and empirical evaluation, it may be necessary to either make it possible for others to replicate the model with the same dataset, or provide access to the model. In general. releasing code and data is often one good way to accomplish this, but reproducibility can also be provided via detailed instructions for how to replicate the results, access to a hosted model (e.g., in the case of a large language model), releasing of a model checkpoint, or other means that are appropriate to the research performed.
        \item While NeurIPS does not require releasing code, the conference does require all submissions to provide some reasonable avenue for reproducibility, which may depend on the nature of the contribution. For example
        \begin{enumerate}
            \item If the contribution is primarily a new algorithm, the paper should make it clear how to reproduce that algorithm.
            \item If the contribution is primarily a new model architecture, the paper should describe the architecture clearly and fully.
            \item If the contribution is a new model (e.g., a large language model), then there should either be a way to access this model for reproducing the results or a way to reproduce the model (e.g., with an open-source dataset or instructions for how to construct the dataset).
            \item We recognize that reproducibility may be tricky in some cases, in which case authors are welcome to describe the particular way they provide for reproducibility. In the case of closed-source models, it may be that access to the model is limited in some way (e.g., to registered users), but it should be possible for other researchers to have some path to reproducing or verifying the results.
        \end{enumerate}
    \end{itemize}

\item {\bf Open access to data and code}
    \item[] Question: Does the paper provide open access to the data and code, with sufficient instructions to faithfully reproduce the main experimental results, as described in supplemental material?
    \item[] Answer: \answerYes{} %
    \item[] Justification: All of our code, as well as logs of queried closed LLMs, are provided in the supplemental materials. All experiments and studies are  reproducible and open source.
    \item[] Guidelines:
    \begin{itemize}
        \item The answer NA means that paper does not include experiments requiring code.
        \item Please see the NeurIPS code and data submission guidelines (\url{https://nips.cc/public/guides/CodeSubmissionPolicy}) for more details.
        \item While we encourage the release of code and data, we understand that this might not be possible, so “No” is an acceptable answer. Papers cannot be rejected simply for not including code, unless this is central to the contribution (e.g., for a new open-source benchmark).
        \item The instructions should contain the exact command and environment needed to run to reproduce the results. See the NeurIPS code and data submission guidelines (\url{https://nips.cc/public/guides/CodeSubmissionPolicy}) for more details.
        \item The authors should provide instructions on data access and preparation, including how to access the raw data, preprocessed data, intermediate data, and generated data, etc.
        \item The authors should provide scripts to reproduce all experimental results for the new proposed method and baselines. If only a subset of experiments are reproducible, they should state which ones are omitted from the script and why.
        \item At submission time, to preserve anonymity, the authors should release anonymized versions (if applicable).
        \item Providing as much information as possible in supplemental material (appended to the paper) is recommended, but including URLs to data and code is permitted.
    \end{itemize}

\item {\bf Experimental setting/details}
    \item[] Question: Does the paper specify all the training and test details (e.g., data splits, hyperparameters, how they were chosen, type of optimizer, etc.) necessary to understand the results?
    \item[] Answer: \answerYes{} %
    \item[] Justification: All experimental hyperparameters, including those pertaining to data selection and model training, are provided in the \cref{app:hyperparams}.
    \item[] Guidelines:
    \begin{itemize}
        \item The answer NA means that the paper does not include experiments.
        \item The experimental setting should be presented in the core of the paper to a level of detail that is necessary to appreciate the results and make sense of them.
        \item The full details can be provided either with the code, in appendix, or as supplemental material.
    \end{itemize}

\item {\bf Experiment statistical significance}
    \item[] Question: Does the paper report error bars suitably and correctly defined or other appropriate information about the statistical significance of the experiments?
    \item[] Answer: \answerYes{} %
    \item[] Justification: We study our method over different seeds. The approximate error is always negligible and reported in \cref{sec:experiments}.
    \item[] Guidelines:
    \begin{itemize}
        \item The answer NA means that the paper does not include experiments.
        \item The authors should answer "Yes" if the results are accompanied by error bars, confidence intervals, or statistical significance tests, at least for the experiments that support the main claims of the paper.
        \item The factors of variability that the error bars are capturing should be clearly stated (for example, train/test split, initialization, random drawing of some parameter, or overall run with given experimental conditions).
        \item The method for calculating the error bars should be explained (closed form formula, call to a library function, bootstrap, etc.)
        \item The assumptions made should be given (e.g., Normally distributed errors).
        \item It should be clear whether the error bar is the standard deviation or the standard error of the mean.
        \item It is OK to report 1-sigma error bars, but one should state it. The authors should preferably report a 2-sigma error bar than state that they have a 96\% CI, if the hypothesis of Normality of errors is not verified.
        \item For asymmetric distributions, the authors should be careful not to show in tables or figures symmetric error bars that would yield results that are out of range (e.g. negative error rates).
        \item If error bars are reported in tables or plots, The authors should explain in the text how they were calculated and reference the corresponding figures or tables in the text.
    \end{itemize}

\item {\bf Experiments compute resources}
    \item[] Question: For each experiment, does the paper provide sufficient information on the computer resources (type of compute workers, memory, time of execution) needed to reproduce the experiments?
    \item[] Answer: \answerYes{} %
    \item[] Justification: The compute resources are described in \cref{app:compute}.
    \item[] Guidelines:
    \begin{itemize}
        \item The answer NA means that the paper does not include experiments.
        \item The paper should indicate the type of compute workers CPU or GPU, internal cluster, or cloud provider, including relevant memory and storage.
        \item The paper should provide the amount of compute required for each of the individual experimental runs as well as estimate the total compute. 
        \item The paper should disclose whether the full research project required more compute than the experiments reported in the paper (e.g., preliminary or failed experiments that didn't make it into the paper). 
    \end{itemize}
    
\item {\bf Code of ethics}
    \item[] Question: Does the research conducted in the paper conform, in every respect, with the NeurIPS Code of Ethics \url{https://neurips.cc/public/EthicsGuidelines}?
    \item[] Answer: \answerYes{} %
    \item[] Justification: We followed the NeurIPS Code of Ethics.
    \item[] Guidelines:
    \begin{itemize}
        \item The answer NA means that the authors have not reviewed the NeurIPS Code of Ethics.
        \item If the authors answer No, they should explain the special circumstances that require a deviation from the Code of Ethics.
        \item The authors should make sure to preserve anonymity (e.g., if there is a special consideration due to laws or regulations in their jurisdiction).
    \end{itemize}

\item {\bf Broader impacts}
    \item[] Question: Does the paper discuss both potential positive societal impacts and negative societal impacts of the work performed?
    \item[] Answer: \answerYes{} %
    \item[] Justification: Social impact is discussed in \cref{sec:impacts}.
    \item[] Guidelines:
    \begin{itemize}
        \item The answer NA means that there is no societal impact of the work performed.
        \item If the authors answer NA or No, they should explain why their work has no societal impact or why the paper does not address societal impact.
        \item Examples of negative societal impacts include potential malicious or unintended uses (e.g., disinformation, generating fake profiles, surveillance), fairness considerations (e.g., deployment of technologies that could make decisions that unfairly impact specific groups), privacy considerations, and security considerations.
        \item The conference expects that many papers will be foundational research and not tied to particular applications, let alone deployments. However, if there is a direct path to any negative applications, the authors should point it out. For example, it is legitimate to point out that an improvement in the quality of generative models could be used to generate deepfakes for disinformation. On the other hand, it is not needed to point out that a generic algorithm for optimizing neural networks could enable people to train models that generate Deepfakes faster.
        \item The authors should consider possible harms that could arise when the technology is being used as intended and functioning correctly, harms that could arise when the technology is being used as intended but gives incorrect results, and harms following from (intentional or unintentional) misuse of the technology.
        \item If there are negative societal impacts, the authors could also discuss possible mitigation strategies (e.g., gated release of models, providing defenses in addition to attacks, mechanisms for monitoring misuse, mechanisms to monitor how a system learns from feedback over time, improving the efficiency and accessibility of ML).
    \end{itemize}
    
\item {\bf Safeguards}
    \item[] Question: Does the paper describe safeguards that have been put in place for responsible release of data or models that have a high risk for misuse (e.g., pretrained language models, image generators, or scraped datasets)?
    \item[] Answer: \answerNA{} %
    \item[] Justification: The method presented in this paper does not pose a high risk for misuse.
    \item[] Guidelines:
    \begin{itemize}
        \item The answer NA means that the paper poses no such risks.
        \item Released models that have a high risk for misuse or dual-use should be released with necessary safeguards to allow for controlled use of the model, for example by requiring that users adhere to usage guidelines or restrictions to access the model or implementing safety filters. 
        \item Datasets that have been scraped from the Internet could pose safety risks. The authors should describe how they avoided releasing unsafe images.
        \item We recognize that providing effective safeguards is challenging, and many papers do not require this, but we encourage authors to take this into account and make a best faith effort.
    \end{itemize}

\item {\bf Licenses for existing assets}
    \item[] Question: Are the creators or original owners of assets (e.g., code, data, models), used in the paper, properly credited and are the license and terms of use explicitly mentioned and properly respected?
    \item[] Answer: \answerYes{} %
    \item[] Justification: We cite all datasets and models used in the paper in \cref{sec:experiments}.
    \item[] Guidelines:
    \begin{itemize}
        \item The answer NA means that the paper does not use existing assets.
        \item The authors should cite the original paper that produced the code package or dataset.
        \item The authors should state which version of the asset is used and, if possible, include a URL.
        \item The name of the license (e.g., CC-BY 4.0) should be included for each asset.
        \item For scraped data from a particular source (e.g., website), the copyright and terms of service of that source should be provided.
        \item If assets are released, the license, copyright information, and terms of use in the package should be provided. For popular datasets, \url{paperswithcode.com/datasets} has curated licenses for some datasets. Their licensing guide can help determine the license of a dataset.
        \item For existing datasets that are re-packaged, both the original license and the license of the derived asset (if it has changed) should be provided.
        \item If this information is not available online, the authors are encouraged to reach out to the asset's creators.
    \end{itemize}

\item {\bf New assets}
    \item[] Question: Are new assets introduced in the paper well documented and is the documentation provided alongside the assets?
    \item[] Answer: \answerNA{} %
    \item[] Justification: We do not release any new assets.
    \item[] Guidelines:
    \begin{itemize}
        \item The answer NA means that the paper does not release new assets.
        \item Researchers should communicate the details of the dataset/code/model as part of their submissions via structured templates. This includes details about training, license, limitations, etc. 
        \item The paper should discuss whether and how consent was obtained from people whose asset is used.
        \item At submission time, remember to anonymize your assets (if applicable). You can either create an anonymized URL or include an anonymized zip file.
    \end{itemize}

\item {\bf Crowdsourcing and research with human subjects}
    \item[] Question: For crowdsourcing experiments and research with human subjects, does the paper include the full text of instructions given to participants and screenshots, if applicable, as well as details about compensation (if any)? 
    \item[] Answer: \answerNA{} %
    \item[] Justification: We do not perform any crowdsourcing or research with human subjects.
    \item[] Guidelines:
    \begin{itemize}
        \item The answer NA means that the paper does not involve crowdsourcing nor research with human subjects.
        \item Including this information in the supplemental material is fine, but if the main contribution of the paper involves human subjects, then as much detail as possible should be included in the main paper. 
        \item According to the NeurIPS Code of Ethics, workers involved in data collection, curation, or other labor should be paid at least the minimum wage in the country of the data collector. 
    \end{itemize}

\item {\bf Institutional review board (IRB) approvals or equivalent for research with human subjects}
    \item[] Question: Does the paper describe potential risks incurred by study participants, whether such risks were disclosed to the subjects, and whether Institutional Review Board (IRB) approvals (or an equivalent approval/review based on the requirements of your country or institution) were obtained?
    \item[] Answer: \answerNA{} %
    \item[] Justification: This paper does not involve crowdsourcing nor research with human subjects.
    \item[] Guidelines:
    \begin{itemize}
        \item The answer NA means that the paper does not involve crowdsourcing nor research with human subjects.
        \item Depending on the country in which research is conducted, IRB approval (or equivalent) may be required for any human subjects research. If you obtained IRB approval, you should clearly state this in the paper. 
        \item We recognize that the procedures for this may vary significantly between institutions and locations, and we expect authors to adhere to the NeurIPS Code of Ethics and the guidelines for their institution. 
        \item For initial submissions, do not include any information that would break anonymity (if applicable), such as the institution conducting the review.
    \end{itemize}

\item {\bf Declaration of LLM usage}
    \item[] Question: Does the paper describe the usage of LLMs if it is an important, original, or non-standard component of the core methods in this research? Note that if the LLM is used only for writing, editing, or formatting purposes and does not impact the core methodology, scientific rigorousness, or originality of the research, declaration is not required.
    \item[] Answer: \answerYes{} %
    \item[] Justification: We describe in \cref{sec:experiments} whenever we use LLMs.
    \item[] Guidelines:
    \begin{itemize}
        \item The answer NA means that the core method development in this research does not involve LLMs as any important, original, or non-standard components.
        \item Please refer to our LLM policy (\url{https://neurips.cc/Conferences/2025/LLM}) for what should or should not be described.
    \end{itemize}

\end{enumerate}

\newpage

\section{Algorithmic Split of Datasets}
\label{app:dataset-split}

We show the split of the BBEH datasets into algorithmic and non-algorithmic datasets in \cref{tab:data-split}. This split of datasets into the two groups is used in analyzing the results in the main body of the paper.

\begin{table}[h]
\centering
\caption{Percent of algorithmic problems in each dataset. We call the datasets with a majority of algorithmic problems the \textit{algorithmic} datasets and the other datasets the \textit{non-algorithmic} datasets.}
\label{tab:data-split}
\begin{tabular}{lc}
\toprule
Dataset & \% Algorithmic \\
\midrule
\multicolumn{2}{l}{\textbf{Non-algorithmic Datasets}} \\
\midrule
Leaf & 0.000 \\
Disambiguation qa & 0.000 \\
Sarc triples & 0.000 \\
Nycc & 0.000 \\
Movie recommendation & 0.000 \\
Hyperbaton & 0.015 \\
Geometric shapes & 0.040 \\
Causal understanding & 0.040 \\
CLEVR & 0.155 \\
Linguini & 0.175 \\
Omnimath-4 & 0.305 \\
Sportqa & 0.345 \\
Omnimath-3 & 0.410 \\
\midrule
\multicolumn{2}{l}{\textbf{Algorithmic Datasets}} \\
\midrule
Clutrr & 0.750 \\
Spatial reasoning & 0.810 \\
Omnimath-2 & 0.835 \\
Buggy tables & 0.940 \\
Web of lies & 0.940 \\
Boardgame qa & 0.955 \\
Object properties & 0.960 \\
Boolean expressions & 0.960 \\
Time arithmetic & 0.970 \\
Word sorting & 0.980 \\
Dyck languages & 0.980 \\
Temporal sequence & 0.990 \\
Omnimath-1 & 0.995 \\
Object counting & 0.995 \\
Zebra puzzles & 0.995 \\
Multistep arithmetic & 1.000 \\
Shuffled objects & 1.000 \\
\bottomrule
\end{tabular}
\end{table}

\section{\ourmethod Programs for \cref{fig:synthesis-failures}}
\label{app:our-programs}

The two corresponding solutions with \ourmethod to the problems shown in \cref{fig:trivial-program} and \cref{fig:opencv} are included in \cref{fig:our-ex1} and \cref{fig:our-ex2} respectively.

\begin{figure}
    \centering
    \begin{minted}{python}
symbols = {'objects': [{'color': 'green', 'material': 'rubber', 'shape': 'cube', 'size': 'large', 'x': 0.1, 'y': 0.4},
    {'color': 'blue', 'material': 'metal', 'shape': 'sphere', 'size': 'large', 'x': 0.4, 'y': 0.3},
    {'color': 'yellow', 'material': 'metal', 'shape': 'cube', 'size': 'large', 'x': 0.6, 'y': 0.3},
    {'color': 'cyan', 'material': 'rubber', 'shape': 'cylinder', 'size': 'small', 'x': 0.7, 'y': 0.3},
    {'color': 'brown', 'material': 'rubber', 'shape': 'sphere', 'size': 'large', 'x': 0.5, 'y': 0.6}]}

def solve(symbols):
  """
  Calculates the number of large objects in front of the cyan object and behind the brown matte ball.

  Args:
    symbols: A dictionary containing information about the objects in the image.

  Returns:
    The number of large objects in front of the cyan object and behind the brown matte ball.
  """

  cyan_x = next((obj["x"] for obj in symbols["objects"] if obj["color"] == "cyan"), None)
  brown_x = next((obj["x"] for obj in symbols["objects"] if obj["color"] == "brown"), None)

  count = 0
  for obj in symbols["objects"]:
    if obj["size"] == "large":
      if obj["x"] < cyan_x and obj["x"] > brown_x:
        count += 1

  return count
    \end{minted}
    \caption{Final symbols and generated program from \ourmethod for the instance shown in \cref{fig:trivial-program}. Executing the code from \ourmethod results in the correct answer of 1.}
    \label{fig:our-ex1}
\end{figure}

\begin{figure}
    \centering
    \small
    \begin{minted}[breaklines]{python}
symbols = {'objects': [
    {'color': 'purple', 'material': 'rubber', 'shape': 'sphere', 'size': 'large', 'x': 0.0},
    {'color': 'purple', 'material': 'metal', 'shape': 'sphere', 'size': 'large', 'x': 1.0},
    {'color': 'blue', 'material': 'rubber', 'shape': 'cube', 'size': 'small', 'x': 2.0},
    {'color': 'gray', 'material': 'rubber', 'shape': 'cylinder', 'size': 'small', 'x': 2.5},
    {'color': 'brown', 'material': 'rubber', 'shape': 'cube', 'size': 'large', 'x': 4.0},
    {'color': 'yellow', 'material': 'rubber', 'shape': 'sphere', 'size': 'large', 'x': 1.5},
    {'color': 'brown', 'material': 'metal', 'shape': 'sphere', 'size': 'small', 'x': 3.5},
    {'color': 'red', 'material': 'metal', 'shape': 'cube', 'size': 'small', 'x': 2.0},
    {'color': 'red', 'material': 'metal', 'shape': 'cube', 'size': 'large', 'x': 5.0}]}
    
def solve(symbols):
    """
    Finds the color of the tiny shiny object behind the big ball to the right of the big metallic thing behind the big brown cube.

    Args:
        symbols (dict): A dictionary containing information about the objects in the image.

    Returns:
        str: The color of the object.
    """

    objects = symbols["objects"]

    # 1. Find the big brown cube
    big_brown_cube = next((obj for obj in objects if obj["shape"] == "cube" and obj["color"] == "brown" and obj["size"] == "large"), None)

    # 2. Find the big metallic thing (red cube) to the right of the big brown cube
    big_metallic_thing = next((obj for obj in objects if obj["shape"] == "cube" and obj["color"] == "red" and obj["size"] == "large" and obj["x"] > big_brown_cube["x"]), None)

    # 3. Find the closest big ball
    closest_big_ball = min((obj for obj in objects if obj["shape"] == "sphere" and obj["size"] == "large"), key=lambda obj: abs(obj["x"] - big_metallic_thing["x"]))

    # 4. Find the tiny shiny object (gold sphere) behind the big ball
    tiny_shiny_object = next((obj for obj in objects if obj["shape"] == "sphere" and obj["material"] == "metal" and obj["size"] == "small" and obj["x"] > closest_big_ball["x"]), None)

    return tiny_shiny_object["color"]
    \end{minted}
    \caption{Final symbols and generated program from \ourmethod for the instance shown in \cref{fig:opencv}. The code returns ``brown'' which is the correct answer.}
    \label{fig:our-ex2}
\end{figure}

\section{Full Results}
\label{app:full-results}

\begin{table}[h]
\centering
\caption{Accuracy and non-trivial-code percentage for o4-mini. Best accuracy per row is bolded.}
\label{tab:o4_mini_2025_04_16_details}
\resizebox{\textwidth}{!}{%
\begin{tabular}{lccccc}
\toprule
Dataset & CoT & PoT & PoT Non-Trivial (\%) & PIPS & PIPS Non-Trivial (\%) \\
\midrule
Buggy tables & 0.275 & 0.512 & 85.6\% & \textbf{0.594} & 96.9\% \\
Temporal sequence & 0.325 & 0.306 & 92.5\% & \textbf{0.519} & 96.2\% \\
Dyck languages & \textbf{0.650} & 0.637 & 8.8\% & 0.606 & 88.8\% \\
Multistep arithmetic & 0.281 & 0.600 & 72.5\% & \textbf{0.631} & 87.5\% \\
Time arithmetic & 0.875 & \textbf{0.900} & 76.2\% & 0.894 & 87.5\% \\
Shuffled objects & 0.081 & 0.150 & 43.1\% & \textbf{0.344} & 71.2\% \\
Web of lies & 0.388 & 0.344 & 68.1\% & \textbf{0.525} & 70.6\% \\
Object counting & 0.850 & 0.812 & 93.1\% & \textbf{0.900} & 70.0\% \\
Zebra puzzles & 0.150 & 0.100 & 62.5\% & \textbf{0.231} & 68.1\% \\
Object properties & 0.144 & 0.219 & 38.8\% & \textbf{0.344} & 65.6\% \\
Boolean expressions & \textbf{0.550} & 0.469 & 24.4\% & 0.419 & 63.7\% \\
Spatial reasoning & 0.463 & 0.506 & 60.6\% & \textbf{0.550} & 55.0\% \\
Word sorting & \textbf{0.806} & 0.775 & 46.2\% & \textbf{0.806} & 53.1\% \\
Omnimath-2 & \textbf{0.812} & 0.706 & 56.9\% & 0.787 & 29.4\% \\
Movie recommendation & \textbf{0.819} & 0.744 & 10.6\% & 0.731 & 18.8\% \\
Omnimath-1 & 0.925 & 0.925 & 78.1\% & \textbf{0.944} & 15.6\% \\
Boardgame qa & \textbf{0.688} & 0.637 & 10.6\% & 0.675 & 13.8\% \\
Hyperbaton & \textbf{0.206} & 0.194 & 34.4\% & 0.188 & 13.8\% \\
Omnimath-3 & \textbf{0.556} & 0.356 & 36.9\% & 0.544 & 11.9\% \\
Omnimath-4 & \textbf{0.662} & 0.419 & 31.2\% & 0.644 & 6.9\% \\
Clutrr & 0.762 & \textbf{0.800} & 1.2\% & 0.762 & 2.5\% \\
Sportqa & \textbf{0.287} & 0.256 & 0.0\% & \textbf{0.287} & 1.9\% \\
Linguini & 0.138 & \textbf{0.175} & 6.2\% & 0.138 & 1.2\% \\
CLEVR & \textbf{0.769} & 0.750 & 6.2\% & \textbf{0.769} & 0.6\% \\
Causal understanding & \textbf{0.581} & 0.550 & 0.6\% & \textbf{0.581} & 0.6\% \\
Leaf & 0.364 & \textbf{0.455} & 0.0\% & 0.364 & 0.0\% \\
Geometric shapes & 0.056 & \textbf{0.119} & 0.0\% & 0.087 & 0.0\% \\
Disambiguation qa & 0.562 & \textbf{0.573} & 0.0\% & 0.562 & 0.0\% \\
Sarc triples & \textbf{0.338} & 0.300 & 0.0\% & \textbf{0.338} & 0.0\% \\
Nycc & \textbf{0.231} & 0.150 & 0.0\% & \textbf{0.231} & 0.0\% \\
\midrule
Harmonic Mean & 0.304 & 0.340 & 0.0\% & 0.397 & 0.1\% \\
\bottomrule
\end{tabular}%
}
\end{table}

\begin{table}[h]
\centering
\caption{Accuracy and non-trivial-code percentage for Gemini-2.0-Flash. Best accuracy per row is bolded.}
\label{tab:gemini_20_flash_details}
\resizebox{\textwidth}{!}{%
\begin{tabular}{lcccccc}
\toprule
Dataset & CoT & PoT & PoT Non-Trivial (\%) & CI & PIPS & PIPS Non-Trivial (\%) \\
\midrule
Shuffled objects & 0.094 & 0.025 & 35.6\% & \textbf{0.537} & 0.188 & 98.1\% \\
Buggy tables & 0.019 & 0.100 & 99.4\% & 0.031 & \textbf{0.188} & 97.5\% \\
Time arithmetic & 0.438 & 0.331 & 82.5\% & 0.312 & \textbf{0.475} & 97.5\% \\
Temporal sequence & 0.006 & 0.006 & 42.5\% & 0.006 & \textbf{0.094} & 96.2\% \\
Multistep arithmetic & \textbf{0.144} & 0.037 & 87.5\% & 0.087 & 0.119 & 90.6\% \\
Dyck languages & \textbf{0.119} & 0.081 & 1.9\% & 0.106 & 0.050 & 90.0\% \\
Boolean expressions & 0.294 & 0.219 & 65.6\% & 0.325 & \textbf{0.456} & 86.2\% \\
Object counting & 0.144 & 0.119 & 98.1\% & 0.181 & \textbf{0.281} & 85.0\% \\
Omnimath-1 & 0.850 & 0.838 & 57.5\% & 0.844 & \textbf{0.869} & 84.4\% \\
Word sorting & 0.287 & 0.525 & 48.8\% & 0.338 & \textbf{0.556} & 73.1\% \\
Object properties & 0.006 & 0.062 & 26.9\% & 0.125 & \textbf{0.163} & 71.9\% \\
Spatial reasoning & 0.231 & \textbf{0.237} & 12.5\% & 0.219 & 0.231 & 70.6\% \\
CLEVR & 0.637 & 0.619 & 26.2\% & 0.669 & \textbf{0.688} & 46.2\% \\
Causal understanding & 0.537 & 0.438 & 1.2\% & \textbf{0.544} & 0.537 & 15.6\% \\
Clutrr & 0.556 & 0.588 & 0.0\% & \textbf{0.662} & 0.506 & 13.8\% \\
Linguini & \textbf{0.144} & 0.113 & 1.9\% & 0.119 & 0.125 & 13.8\% \\
Boardgame qa & \textbf{0.463} & 0.419 & 5.0\% & 0.394 & \textbf{0.463} & 11.9\% \\
Zebra puzzles & \textbf{0.300} & 0.256 & 73.8\% & 0.131 & 0.275 & 10.0\% \\
Geometric shapes & 0.312 & 0.269 & 0.6\% & \textbf{0.388} & 0.300 & 9.4\% \\
Sportqa & 0.200 & 0.244 & 1.9\% & \textbf{0.269} & 0.194 & 6.9\% \\
Hyperbaton & \textbf{0.031} & 0.019 & 46.2\% & \textbf{0.031} & 0.025 & 5.6\% \\
Web of lies & \textbf{0.219} & 0.206 & 3.1\% & 0.188 & \textbf{0.219} & 2.5\% \\
Movie recommendation & \textbf{0.581} & 0.562 & 0.0\% & 0.556 & 0.569 & 2.5\% \\
Disambiguation qa & 0.448 & 0.417 & 0.0\% & \textbf{0.479} & 0.448 & 1.0\% \\
Leaf & 0.602 & \textbf{0.636} & 0.0\% & 0.102 & 0.602 & 0.0\% \\
Sarc triples & \textbf{0.375} & 0.369 & 0.0\% & 0.344 & \textbf{0.375} & 0.0\% \\
Nycc & 0.106 & \textbf{0.131} & 0.0\% & 0.113 & 0.106 & 0.0\% \\
Omnimath-2 & \textbf{0.544} & 0.463 & 46.9\% & \textbf{0.544} & \textbf{0.544} & 0.0\% \\
Omnimath-3 & 0.269 & 0.194 & 15.6\% & \textbf{0.275} & 0.269 & 0.0\% \\
Omnimath-4 & 0.312 & 0.244 & 16.2\% & \textbf{0.319} & 0.312 & 0.0\% \\
\midrule
Harmonic Mean & 0.107 & 0.115 & 0.0\% & 0.134 & 0.201 & 0.0\% \\
\bottomrule
\end{tabular}%
}
\end{table}

\begin{table}[h]
\centering
\caption{Accuracy and non-trivial-code percentage for gpt-4.1-mini-2025-04-14. Best accuracy per row is bolded.}
\label{tab:gpt_41_mini_2025_04_14_details}
\resizebox{\textwidth}{!}{%
\begin{tabular}{lccccc}
\toprule
Dataset & CoT & PoT & PoT Non-Trivial (\%) & PIPS & PIPS Non-Trivial (\%) \\
\midrule
Time arithmetic & 0.588 & \textbf{0.706} & 93.1\% & 0.463 & 98.8\% \\
Buggy tables & 0.075 & \textbf{0.481} & 100.0\% & 0.406 & 98.1\% \\
Multistep arithmetic & 0.275 & 0.394 & 86.2\% & \textbf{0.494} & 96.2\% \\
Dyck languages & 0.150 & 0.175 & 6.9\% & \textbf{0.506} & 95.0\% \\
Shuffled objects & 0.119 & 0.256 & 68.8\% & \textbf{0.294} & 95.0\% \\
Omnimath-1 & \textbf{0.894} & 0.875 & 76.2\% & 0.875 & 89.4\% \\
Word sorting & \textbf{0.681} & 0.600 & 73.1\% & 0.656 & 85.6\% \\
Object counting & 0.263 & \textbf{0.331} & 66.2\% & 0.287 & 83.1\% \\
Temporal sequence & 0.250 & \textbf{0.356} & 96.2\% & 0.275 & 82.5\% \\
Boolean expressions & 0.294 & 0.356 & 11.2\% & \textbf{0.469} & 79.4\% \\
Spatial reasoning & 0.181 & \textbf{0.394} & 48.1\% & 0.362 & 68.1\% \\
Object properties & 0.025 & \textbf{0.237} & 67.5\% & 0.175 & 60.0\% \\
Omnimath-2 & \textbf{0.569} & \textbf{0.569} & 63.1\% & 0.556 & 59.4\% \\
Web of lies & \textbf{0.362} & 0.300 & 36.2\% & 0.219 & 58.8\% \\
CLEVR & \textbf{0.719} & 0.669 & 8.1\% & 0.700 & 53.8\% \\
Zebra puzzles & \textbf{0.194} & 0.150 & 36.2\% & 0.188 & 49.4\% \\
Clutrr & \textbf{0.662} & 0.562 & 0.6\% & 0.613 & 21.9\% \\
Omnimath-3 & \textbf{0.338} & 0.244 & 35.6\% & 0.325 & 16.2\% \\
Geometric shapes & \textbf{0.344} & 0.294 & 16.9\% & 0.319 & 15.0\% \\
Boardgame qa & 0.512 & 0.500 & 81.2\% & \textbf{0.519} & 11.2\% \\
Omnimath-4 & \textbf{0.463} & 0.312 & 28.7\% & 0.431 & 10.0\% \\
Sportqa & 0.169 & \textbf{0.244} & 2.5\% & 0.200 & 9.4\% \\
Hyperbaton & \textbf{0.087} & 0.062 & 4.4\% & 0.075 & 7.5\% \\
Causal understanding & \textbf{0.562} & \textbf{0.562} & 1.2\% & 0.550 & 5.6\% \\
Linguini & 0.094 & \textbf{0.144} & 1.2\% & 0.094 & 3.1\% \\
Movie recommendation & \textbf{0.606} & 0.475 & 8.8\% & 0.594 & 1.9\% \\
Leaf & 0.341 & \textbf{0.409} & 0.0\% & 0.341 & 0.0\% \\
Disambiguation qa & \textbf{0.552} & 0.500 & 1.0\% & \textbf{0.552} & 0.0\% \\
Sarc triples & 0.287 & \textbf{0.331} & 10.0\% & 0.287 & 0.0\% \\
Nycc & \textbf{0.188} & 0.150 & 15.0\% & \textbf{0.188} & 0.0\% \\
\midrule
Harmonic Mean & 0.211 & 0.297 & 0.3\% & 0.305 & 0.1\% \\
\bottomrule
\end{tabular}%
}
\end{table}

\begin{table}[h]
\centering
\caption{Accuracy for Qwen3-235B-A22B over all BBEH tasks. Best accuracy per row is bolded.}
\label{tab:qwen3_235b_details}
\begin{tabular}{lccc}
\toprule
Dataset & CoT & PoT & PIPS \\
\midrule
Word sorting & \textbf{0.738} & 0.688 & 0.706 \\
Dyck languages & 0.438 & 0.200 & \textbf{0.456} \\
Object counting & \textbf{0.644} & 0.519 & 0.613 \\
Object properties & 0.319 & \textbf{0.475} & 0.250 \\
Boardgame qa & 0.744 & \textbf{0.838} & 0.744 \\
Boolean expressions & 0.600 & 0.362 & \textbf{0.631} \\
Buggy tables & 0.181 & \textbf{0.362} & 0.312 \\
Spatial reasoning & 0.512 & 0.506 & \textbf{0.519} \\
Multistep arithmetic & \textbf{0.550} & 0.544 & 0.275 \\
Geometric shapes & \textbf{0.237} & 0.200 & \textbf{0.237} \\
Temporal sequence & 0.294 & 0.331 & \textbf{0.350} \\
Disambiguation qa & \textbf{0.625} & 0.573 & \textbf{0.625} \\
Causal understanding & \textbf{0.569} & 0.456 & \textbf{0.569} \\
Time arithmetic & 0.613 & \textbf{0.750} & 0.700 \\
Web of lies & \textbf{0.812} & 0.631 & \textbf{0.812} \\
Sarc triples & \textbf{0.281} & 0.194 & \textbf{0.281} \\
Hyperbaton & \textbf{0.287} & \textbf{0.287} & \textbf{0.287} \\
Nycc & 0.156 & \textbf{0.163} & 0.156 \\
Sportqa & \textbf{0.256} & 0.150 & \textbf{0.256} \\
Linguini & \textbf{0.175} & 0.150 & \textbf{0.175} \\
Movie recommendation & \textbf{0.738} & 0.719 & \textbf{0.738} \\
Shuffled objects & 0.144 & 0.056 & \textbf{0.569} \\
Zebra puzzles & \textbf{0.700} & 0.569 & \textbf{0.700} \\
\midrule
Harmonic Mean & 0.355 & 0.289 & \textbf{0.386} \\
\bottomrule
\end{tabular}
\end{table}

\begin{table}[ht]
\centering
\caption{Harmonic Mean Accuracy (All Datasets)}
\begin{tabular}{lccc}
\toprule
Method & gpt-4.1-mini & Gemini-2.0-Flash & o4-mini \\
\midrule
CoT   & 0.211 & 0.107 & 0.304 \\
PoT   & 0.297 & 0.115 & 0.340 \\
PIPS  & 0.305 & 0.201 & 0.397 \\
CI    & 0.000 & 0.134 & 0.000 \\
\bottomrule
\end{tabular}
\end{table}

\begin{table}[ht]
\centering
\caption{Harmonic Mean Accuracy (Algorithmic Datasets)}
\begin{tabular}{lccc}
\toprule
Method & gpt-4.1-mini & Gemini-2.0-Flash & o4-mini \\
\midrule
CoT   & 0.187 & 0.079 & 0.339 \\
PoT   & 0.357 & 0.094 & 0.389 \\
PIPS  & 0.369 & 0.217 & 0.548 \\
CI    & - & 0.112 & - \\
\bottomrule
\end{tabular}
\end{table}

\begin{table}[ht]
\centering
\caption{Harmonic Mean Accuracy (Non-Algorithmic Datasets)}
\begin{tabular}{lccc}
\toprule
Method & gpt-4.1-mini & Gemini-2.0-Flash & o4-mini \\
\midrule
CoT   & 0.254 & 0.199 & 0.267 \\
PoT   & 0.244 & 0.166 & 0.291 \\
PIPS  & 0.248 & 0.184 & 0.292 \\
CI    & - & 0.181 & - \\
\bottomrule
\end{tabular}
\end{table}

The full results over all 30 datasets for Gemini-2.0-Flash, GPT-4.1-mini, and o4-mini and included in \cref{tab:gemini_20_flash_details}, \cref{tab:gpt_41_mini_2025_04_14_details}, and \cref{tab:o4_mini_2025_04_16_details} respectively. Results for an open-weights model, Qwen3-235B-A22B, for BBEH tasks is included in \cref{tab:qwen3_235b_details}.

\section{Additional Baselines}
\label{app:more-baselines}

In this section, we compare \ourmethod with an iterative refinement version of PoT, CodeAct \citep{wang2024executable}, and Buffer-of-Thoughts (BoT) \citep{yang2024buffer}.
The method PoT-retries refers to our modified version of PoT which regenerates the program if the produced program results in an execution error. We use the CodeAct implementation provided with the smolagents library from huggingface. Finally, we use the code for BoT from \citet{yang2024buffer} for this baseline and we additionally fixed several bugs which previously resulted in a high failure rate.

Results for PoT-retries, CodeAct, BoT, and \ourmethod are included in \cref{tab:more-baselines}. We show results using Gemini-2.0-Flash over all BBEH tasks since BoT does not support the multimodal tasks.

\begin{table}[h]
    \centering
    \caption{Harmonic mean accuracy computed over BBEH tasks for additional agentic baselines compared to \ourmethod on Gemini-2.0-Flash. The highest accuracy is bolded.}
    \label{tab:more-baselines}
    \begin{tabular}{lr}
    \toprule
    Method & HMean Accuracy \\
    \midrule
         PoT & 0.095\\
         PoT-retries &  0.098\\
         CodeAct & 0.040\\
         BoT & 0.027\\
         \ourmethod & \textbf{0.171}\\
    \bottomrule
    \end{tabular}
\end{table}

\section{Reasoning Costs}
\label{app:cost}

In addition to comparing the costs of different methods for using code for solving challenging reasoning problems, we also compare the cost in terms of number of tokens and dollar cost. \Cref{tab:token-cost} shows average input, output, and dollar cost averaged over all 30 datasets for Gemini-2.0-flash. We find that the iterative approaches (CodeAct and Buffer of Thoughts) increase cost by more than 10X compared to PoT or CoT, but \ourmethod{} is overall only 3-4X more expensive than CoT or PoT while achieving much greater accuracy.

\begin{table}[h]
\centering
\caption{Comparison of average token usage and cost across methods.}
\label{tab:token-cost}
\begin{tabular}{lrrr}
\toprule
\textbf{Method} & \textbf{Avg. Input Tokens} & \textbf{Avg. Output Tokens} & \textbf{Cost (USD)} \\
\midrule
PoT & 1{,}115.96 & 1{,}333.98 & \$0.0006 \\
CoT & 1{,}099.77 & 1{,}475.87 & \$0.0007 \\
CodeAct & 80{,}137.92 & 4{,}023.36 & \$0.0096 \\
Buffer of Thoughts & 340{,}927.19 & 123{,}655.35 & \$0.0835 \\
\ourmethod & 11{,}839.09 & 2{,}805.78 & \$0.0023 \\
\bottomrule
\end{tabular}
\end{table}

\section{Non-Trivial Program Synthesis on Non-Algorithmic Problems}
\label{app:pips-nonalgo}
We find that encouraging non-trivial programs for non-algorithmic problems leads to reduced performance. For instance, CoT with Gemini-2.0-Flash results in a harmonic mean accuracy over all non-algorithmic datasets (as listed in \cref{tab:data-split}) of 0.199 while PoT results in a lower value of 0.166 and our method without switching results in 0.151. Producing non-trivial programs for non-algorithmic problems which shouldn't be solved via code in the first place, harms performance. Therefore, a high performing general reasoning system needs to avoid program synthesis in such cases.

\section{Program Evaluation Criteria}
\label{app:eval-criteria}
The eight criteria we use to evaluate code within \ourmethod are included below. The input dependence criteria is meant to catch when the program is trivial, the proper output criteria catches cases where the program does not output the answer in the correct format, and we also include the symbol extraction issues criteria to find issues during the first step of symbol extraction.
\begin{itemize}
    \item Input dependence: Does the code use the input symbols to compute the answer?
    \item Valid return: Does the code avoid returning \texttt{None} unless it is the correct answer?
    \item Proper output: Does the code return (not print) the correct answer in the expected format?
    \item No example usage: Does the code omit example calls or usage?
    \item Simplifiability: Could the solution be implemented in a simpler way?
    \item Correctness bugs: Are there any bugs affecting correctness?
    \item Symbol extraction issues: Are there any problems with the extracted input symbols?
    \item Sanity check: Does the output pass a basic sanity check?
\end{itemize}

\section{Switch Analysis}
In this section, we go in-depth on the CoT vs. program synthesis switch design. First we discuss some additional evaluations involving evaluating the generalizability of the trained switch as well as evaluating a zero-shot switch.

\subsection{Switch Ablations}
\label{app:switch-ablations}

We perform ablations for the switch in \ourmethod and show the results in \cref{tab:switch-ablations}. First, we show that if there is no calibration data available to train the switch, it can be used in a zero-shot manner and still be highly performant. The zero-shot switch uses only the last of the ten questions as the final classifier value so that no training is  required.

To validate that training a classifier on any data (even from a different dataset) can still be useful, we perform a leave-one-dataset-out evaluation. This involves training the switch on all but one dataset and then evaluating \ourmethod over only the left out dataset and averaging performance over all datasets as the left out one. As shown in \cref{tab:switch-ablations}, this also performs nearly as well as \ourmethod where the switch is trained over calibration set of data sampled from all datasets.

\begin{table}[h]
    \centering
    \caption{Ablations for the switch in \ourmethod compared using harmonic mean accuracy over all 30 datasets. The ZS-Switch uses only the model output from the final of the ten criteria as the final switch decision the leave-one-out (LOO) setting trains the switch over all but one dataset and then evaluates on the left out dataset and averages over all datasets being the left out one.}
    \label{tab:switch-ablations}
    \begin{tabular}{lrrr}
    \toprule
         Models & \ourmethod w. ZS-Switch & \ourmethod w. LOO & \ourmethod \\
         \midrule
         o4-mini	& 0.389	& 0.393	& 0.397\\
        Gemini-2.0-Flash & 0.199 & 0.180 & 0.208\\
        GPT-4.1-mini & 0.259 & 0.302 & 0.305\\
         \bottomrule
    \end{tabular}
\end{table}

\subsection{Switch Criteria}
\label{app:switch-criteria}
The full prompt for the switch is provided in \cref{app:prompts}, but we provide the 10 criteria we use within the prompt below. The criteria for determining if an instance should be solved directly via CoT or by program synthesis are the following:
\begin{enumerate}
    \item Simple formalizability: Likelihood that the solution can be easily expressed as simple, deterministic code.
    \item Straightforward executability: Likelihood that a first code attempt runs correctly without debugging.
    \item Robust systematic search: Likelihood that systematic code (e.g., brute-force, recursion) reliably solves the problem.
    \item Manageable state representation: Likelihood that all necessary variables and concepts can be cleanly represented in code.
    \item Structured knowledge encoding: Likelihood that required background knowledge can be encoded as rules or data.
    \item Hallucination risk reduction: Likelihood that code avoids fabricated steps better than chain-of-thought reasoning.
    \item Arithmetic and data processing advantage: Likelihood that code handles arithmetic or data processing more reliably.
    \item Branching and case handling advantage: Likelihood that code handles special cases or branching logic more systematically.
    \item Algorithmic reliability over heuristics: Likelihood that a deterministic algorithm outperforms intuitive reasoning.
    \item Overall comparative success: Likelihood that code yields a more reliable solution than chain-of-thought reasoning.
\end{enumerate}
Our full prompt asks the LLM itself to quantify each of these criteria and then we build a simple logistic classifier based on the LLM's own judgements to determine when to use program synthesis.

\subsection{Examples of Switch Decisions}
\label{app:switch-decisions}

\begin{figure}[ht]
\centering
\noindent
\begin{minipage}[t]{0.49\textwidth}
\begin{tcolorbox}[title=Q1, fontupper=\scriptsize]
You are an expert in word sorting. You will be provided with a list of words and the steps taken, in the form of thoughts, to arrange them in alphabetical order.
Your job is to identify the first step that was a mistake in reasoning about the order of the words. This can be misunderstanding the order of
the alphabet or getting the word or letter wrong or missing a word entirely.
Q: Sort the following words alphabetically: List: electron amputate morphism tire scald lawmen putdown fishpond expositor yosemite gourmet transplant noon structure protagonist arab agree daisy delta eskimo
Thought 1: I should start by looking at the first letter of the words in the list. The first letter: "electron": "e" (5). "amputate": "a" (1). "morphism": "m" (13). "tire": "t" (20). "scald": "s" (19). "lawmen": "l" (12). "putdown": "p" (16). "fishpond": "f" (6). "expositor": "e" (5). "yosemite": "y" (25). "gourmet": "g" (7). "transplant": "t" (20). "noon": "n" (14). "structure": "s" (19). "protagonist": "p" (16). "arab": "a" (1). "agree": "a" (1). "daisy": "d" (4). "delta": "d" (4). "eskimo": "e" (5).

...

Thought 15: I have now sorted all the words. The answer is agree amputate arab delta daisy electron expositor eskimo fishpond gourmet lawmen morphism noon protagonist putdown scald structure tire transplant yosemite
Q: Is there a mistake in this sequence? Write "No" if there are no mistakes, or the number N if there is a mistake in Thought N.
\end{tcolorbox}
\end{minipage}
\hfill
\begin{minipage}[t]{0.49\textwidth}
\begin{tcolorbox}[title=Q2, fontupper=\scriptsize]
Consider a new alphabet whose letters have the same order as the English alphabet, except that c and m are the last two letters.
Sort the following words with the new alphabet and separate them with comma:
medea, oversimplifications, clonic, chaplin, kennan, postpone, squabble, ipsilateral, misunderstandings, ussr, canal, modifications, referring, counterrevolutionaries, pyridine, cameroon, avalanche, rationalizations, fortran, cram, coachman
\end{tcolorbox}
\end{minipage}
\caption{Two questions from the BBEH Word Sorting task where the first question asks for determining which thought in a model's CoT is incorrect and the second question asks for sorting a list of words. The switch for these problems chooses to answer Q1 with CoT for all models while answer Q2 with program synthesis for all models.}
\label{fig:switch1}
\end{figure}

\begin{figure}[ht]
\centering
\begin{tcolorbox}[title=Q1, fontupper=\scriptsize]
A collection $\mathcal{S}$ of 10000 points is formed by picking each point uniformly at random inside a circle of radius 1. Let $N$ be the expected number of points of $\mathcal{S}$ which are vertices of the convex hull of the $\mathcal{S}$. (The convex hull is the smallest convex polygon containing every point of $\mathcal{S}$.) Estimate $N$.
\end{tcolorbox}
\caption{A question from Omnimath-2 where both Gemini-2.0-Flash and GPT-4.1-mini switch to CoT to answer while o4-mini chooses program synthesis.}
\label{fig:switch2}
\end{figure}

We show two questions from BBEH Word Sorting in \cref{fig:switch1} and the corresponding switch decisions for the different models. We also show a question from Omnimath-2 in \cref{fig:switch2} where the different CoT/Synthesis switch decisions are made for different models.

\subsection{Question Analysis}
\label{app:switch-questions}

We include the logistic regression weights for each of the ten questions in \cref{app:switch-criteria} in \cref{tab:lr_coefficients}. The most important questions for the switch actually vary significantly between models. For Gemini-2.0-Flash, question 5 and 6 are the most important while for GPT-4.1-mini it is question 8 and 5. Finally, for o4-mini, questions 2 and 1 are the most important. Interestingly, we see that questions 5, 6, and 8 which are important for the non-reasoning models are related to the ability of the model to produce error-free code and avoid tedious steps using code while the reasoning model relies most on the criteria which concerns traditional problem algorithmicity rather than model capability.

\begin{table}[h]
\centering
\small
\caption{Logistic regression coefficients for the switch for each of the three models.}
\label{tab:lr_coefficients}
\begin{tabular}{lcccccccccc}
\toprule
Model & Q1 & Q2 & Q3 & Q4 & Q5 & Q6 & Q7 & Q8 & Q9 & Q10 \\
\midrule
Gemini-2.0-Flash & 0.14 & 0.03 & 0.12 & 0.15 & 0.21 & -0.21 & 0.18 & -0.09 & 0.03 & 0.10 \\
gpt-4.1-mini & 0.40 & 0.42 & 0.20 & 0.28 & 0.22 & 0.15 & -0.02 & 0.01 & 0.21 & 0.21 \\
o4-mini & 0.22 & 0.04 & 0.16 & 0.24 & 0.27 & -0.05 & 0.12 & 0.35 & 0.18 & 0.24 \\
\bottomrule
\end{tabular}
\end{table}

\section{Prompts}
\label{app:prompts}

All prompts used in our evaluation and method are included below.

The prompt used to create the LLM-based classifier for question algorithmicity is the following.
\begin{promptbox}[Algorithmic Question Evaluation]
You will determine whether a given target question can be definitively solved by writing a Python program (algorithmic) or if it necessitates another form of reasoning (non-algorithmic). A Python solution may import standard libraries, but cannot simply invoke external services, APIs, or LLMs.
If the input contains images, an algorithmic solution may use information manually extracted without needing to interpret the image itself.

Evaluate the target question carefully against the following criteria. Answer each sub-question rigorously with a binary response (1 for yes, 0 for no), ensuring a high threshold for certainty:

1. Does the problem have explicitly defined inputs and outputs, such that identical inputs always yield identical outputs?
2. Are there explicit, clearly stated rules, formulas, algorithms, or known deterministic procedures available for solving this problem?
3. Does solving this problem strictly require exact computation (no approximations, intuition, or interpretation)?
4. Can this problem be fully formalized in clear mathematical, logical, or structured computational terms without ambiguity?
5. Can the solution method be decomposed into a finite, clear, and unambiguous sequence of computational steps?
6. Is there a universally recognized and objective standard for verifying the correctness of the solution?
7. Does solving the problem inherently involve repetitive or iterative computations clearly suitable for automation?
8. Are the inputs structured, quantifiable, and inherently suited to algorithmic manipulation?
9. Does this problem clearly match or closely resemble a known, standardized computational task or problem type?
10. Is absolute correctness required (i.e., no margin for error or subjective interpretation)?

After thoroughly reasoning through these sub-questions, append a final determination as the 11th element:
- Output 1 if and only if all or nearly all (at least 8 out of 10) answers are clearly 1, indicating the problem is definitively algorithmic.
- Otherwise, output 0, indicating the problem requires non-algorithmic reasoning.

IMPORTANT:
- Before providing the binary list, explicitly reason through each criterion carefully and thoroughly, clearly justifying your decisions. If uncertain or ambiguous about any criterion, default to 0.
- Provide your final answer explicitly as an 11-element binary list (ten answers plus the final determination).
- Under no circumstances should you attempt to answer the actual target question itself.

TARGET QUESTION:
\end{promptbox}

The prompt used to extract the ten criteria for building our instance-level CoT or program synthesis switch is provided next.
\begin{promptbox}[CoT or Synthesis Switch Criteria]
You will self-reflect to estimate whether you are more likely to correctly solve a given target question by writing executable Python code or by using chain-of-thought (natural-language) reasoning.

**IMPORTANT:**
- This is a hypothetical evaluation.
- **You must NOT attempt to answer, solve, write code, or reason through the target question yet.**
- Instead, you must reflect carefully and conservatively on your expected ability if you were to attempt solving the question through either method.

Solution Expectations:
- You may assume standard library modules are allowed for code.
- You may NOT call external services, APIs, databases, or other LLMs.
- The code must be self-contained and executable without internet access.
- Chain-of-thought reasoning must be clear, logically sound, and internally verifiable without external tools.

**CRITICAL GUIDANCE:**
- **Be cautious, not optimistic.**  
  Overestimating your capabilities will lead to choosing a method you cannot successfully complete.
- **If you feel any uncertainty, complexity, or ambiguity, lower your probability accordingly.**
- **Assume that even small mistakes can cause failure** when writing code or reasoning through complex tasks.
- **Use conservative estimates.**
- If unsure between two options, **prefer lower probabilities rather than guessing high**.

Here are the self-reflection sub-questions you must answer hypothetically:

1. **Simple Formalizability** - *What is the probability that the full solution can be easily and directly expressed as simple, deterministic code, without needing complex transformations or deep insight?*

2. **Straightforward Executability** - *What is the probability that a first attempt at writing code would execute correctly without needing debugging, even if the problem has subtle or complex aspects?*

3. **Robust Systematic Search** - *What is the probability that coding a systematic method (like brute-force search or recursion) would reliably find the correct answer, without missing hidden constraints or introducing edge-case errors?*

4. **Manageable State Representation** - *What is the probability that all intermediate concepts, variables, and conditions can be simply and explicitly represented in code, without requiring difficult or error-prone state tracking?*

5. **Structured Knowledge Encoding** - *What is the probability that all required background knowledge can be neatly encoded in code (e.g., as rules, formulas, or data), rather than needing flexible, intuitive understanding better suited to reasoning?*

6. **Hallucination Risk Reduction** - *What is the probability that code execution would more reliably avoid fabricated steps or unwarranted assumptions compared to chain-of-thought reasoning?*

7. **Arithmetic and Data Processing Advantage** - *What is the probability that the problem requires extensive or error-prone arithmetic/data handling that code could perform perfectly, but that chain-of-thought would likely fumble?*

8. **Branching and Case Handling Advantage** - *What is the probability that the solution involves many branching conditions, special cases, or exceptions that code can handle systematically but chain-of-thought might overlook?*

9. **Algorithmic Reliability Over Heuristics** - *What is the probability that following a deterministic algorithm in code would reach the correct answer more reliably than relying on intuitive or heuristic chain-of-thought reasoning?*

10. **Overall Comparative Success** - *Considering all factors, what is the probability that code will ultimately produce a correct solution more reliably than chain-of-thought reasoning for this question?*

After thoroughly reasoning through each criterion:

- Output a single list of 10 probability scores (each between 0 and 1) as your FINAL ANSWER, in order:
  - Scores 1-10 correspond to the ten sub-questions above.

**Additional Instructions:**
- Explicitly reason through each criterion carefully before giving a probability.
- If uncertain or if the problem seems complex, favor lower probabilities to reflect the difficulty.
- Make sure to put only the list after FINAL ANSWER.
- **Under no circumstances should you write, sketch, pseudocode, or attempt any part of the solution itself during this reflection phase.**

TARGET QUESTION: 
\end{promptbox}

The prompt for generating the first program with \ourmethod in iteration one is the following.
\begin{promptbox}[PIPS Code Generator (Iteration 0)]
You will be given a question and you must answer it by extracting relevant symbols in JSON format and then writing a Python program to calculate the final answer.

You MUST always plan extensively before outputting any symbols or code.

You MUST iterate and keep going until the problem is solved.

# Workflow

## Problem Solving Steps
1. First extract relevant information from the input as JSON. Try to represent the relevant information in as much of a structured format as possible to help with further reasoning/processing.
2. Using the information extracted, determine a reasonable approach to solving the problem using code, such that executing the code will return the final answer.
3. Write a Python program to calculate and return the final answer. Use comments to explain the structure of the code and do not use a main() function.
The JSON must be enclosed in a markdown code block and the Python function must be in a separate markdown code block and be called `solve` and accept a single input called `symbols` representing the JSON information extracted. Do not include any `if __name__ == "__main__"` statement and you can assume the JSON will be loaded into the variable called `symbols` by the user.
The Python code should not just return the answer or perform all reasoning in comments and instead leverage the code itself to perform the reasoning.
Be careful that the code returns the answer as expected by the question, for instance, if the question is multiple choice, the code must return the choice as described in the question.
Be sure to always output a JSON code block and a Python code block.
\end{promptbox}

The prompt used to generate subsequent code solutions with \ourmethod by leveraging the evaluator output is shown below.
\begin{promptbox}[PIPS Code Generator (Iteration > 0)]
Please fix the issues with the code and symbols or output "FINISHED".
The following is the result of evaluating the above code with the extracted symbols.
```
Return value: {output}
Standard output: {stdout}
Exceptions: {err}
```

The following is the summary of issues found with the code or the extracted symbols by another model:
```
{checker_output}
```

If there are any issues which impact the correctness of the answer, please output code which does not have the issues. Before outputting any code, plan how the code will solve the problem and avoid the issues.
If stuck, try outputting different code to solve the problem in a different way.
You may also revise the extracted symbols. To do this, output the revised symbols in a JSON code block. Only include information in the JSON which is present in the original input to keep the code grounded in the specific problem. Some examples of symbol revisions are changing the names of certain symbols, providing further granularity, and adding information which was originally missed.
If everything is correct, output the word "FINISHED" and nothing else.
\end{promptbox}

Finally, the prompt for the code evaluator is shown below.
\begin{promptbox}[PIPS Evaluator]
You will be given a question and a code solution and you must judge the quality of the code for solving the problem.
                           
Look for any of the following issues in the code:
- The code should be input dependent, meaning it should use the input symbols to compute the answer. It is OK for the code to be specialized to the input (i.e. the reasoning itself may be hardcoded, like a decision tree where the branches are hardcoded).
- The code should not return None unless "None" is the correct answer.
- The code should return the answer, not just print it. If the question asks for a multiple choice answer, the code should return the choice as described in the question.
- There should not be any example usage of the code.
- If there is a simpler way to solve the problem, please describe it.
- If there are any clear bugs in the code which impact the correctness of the answer, please describe them.
- If there are any issues with the extracted symbols, please describe them as well, but separate these issues from the issues with the code.
- If it is possible to sanity check the output of the code, please do so and describe if there are any obvious issues with the output and how the code could be fixed to avoid these issues.

After analyzing the code in depth, output a concrete and concise summary of the issues that are present, do not include any code examples. Please order the issues by impact on answer correctness.
The following are extracted symbols from the question in JSON format followed by a Python program which takes the JSON as an argument called `symbols` and computes the answer.
```json
{json_str}
```

```python
{code_str}
```

Code execution result:
```
Return value: {output}
Standard output: {stdout}
Exceptions: {err}
```

Output a concrete and concise summary of only the issues that are present, do not include any code examples.
\end{promptbox}

\section{Hyperparameters}
\label{app:hyperparams}

For all models we used a temperature of 0.0. For \ourmethod, we used a maximum of 30 iterations for all models.

\section{Compute Resources}
\label{app:compute}

For most experiments we rely on API model access. All experiments cost \$300 for Gemini-2.0-Flash, \$70 GPT-4.1-mini, and \$700 for o4-mini (medium). Other experiments were run on a server with 96 Intel(R) Xeon(R) Gold 5318Y CPUs @ 2.10GHz with 1TB of system RAM. The server also had 10x NVIDIA A100 80GB GPUs which were only used for local testing of open-weights models.

\section{Broader Impacts}
\label{sec:impacts}
\ourmethod offers significant potential benefits, including enhanced AI reliability, trust, transparency, and the democratization of advanced problem-solving. However, like most systems built from powerful foundation models, it also presents important considerations. These include the potential for bias propagation from underlying models, challenges in ensuring the complexity and robustness of dynamically generated programs, and concerns regarding misuse. Continued research and development must prioritize robust safeguards, fairness, transparency, and responsible deployment practices to harness the benefits while mitigating these potential negative impacts.

\section{Connection to Transductive Learning}
Instance-wise Program Synthesis can be connected to transductive learning \citep{vapnik1999nature} where a function is learned to map from specific function inputs to specific outputs. The general philosophy is that one should not try to solve the general problem when the specific case is all one needs.

We are given a labeled training set $D_L=\{(x_i, y_i)\}_{i=1}^m \in \mathcal{R}\times\mathcal{Y}$ where $\mathcal{R}$ is the space of symbolic input and $\mathcal{Y}$ is the output space, and an unlabeled test sample $x\in\mathcal{R}$. Our goal is to find a program $p: \mathcal{R}\rightarrow\mathcal{Y}$ which has a low error on the given sample, so $p(x)=y$. In contrast, inductive program synthesis tries to find a program $p$ which \textit{generalizes}, so it should achieve a low error on samples from the same distribution as $D_L$. To perform transductive program synthesis, we rely on minimizing an auxiliary ``regularization'' term, $\Omega$, over the test sample, representing a form of test-time computation:

\begin{align*}
    \hat{p} = \arg\min_p \left\{ \frac{1}{m} \sum_{i=1}^m \ell(p(x_i), y_i) + \Omega(p; x_1, \dots, x_m, x)\right\}.
\end{align*}

The regularization term can be implemented using probabilistic priors (e.g. the likelihood under some language model) or more complex heuristics involving static/dynamic code analysis. With powerful foundation models that can perform zero-shot inference, we can even perform this type of learning without any training set.

\paragraph{Neuro-symbolic Instance-level Synthesis}
The above problem definition is specifically for problems formulated for program synthesis, meaning the inputs are expressed in symbolic form. Can we apply such a method for problems expressed in natural language or even using images? To convert general problems into a form which is conducive to program execution, we use a Neuro-Symbolic framework, specifically a Prompt-Symbolic approach.

We now assume our samples come from an arbitrary raw input space, such as natural language, images, and other modalities. Using the traditional neuro-symbolic framework, we first extract symbols from the raw input and then pass these symbols to a program. Let $C: \mathcal{X}\rightarrow\mathcal{R}$ be a mapping from raw data to symbols. Now we adapt the above formalism as the following:

\begin{align*}
    \hat{p}, \hat{C} = \arg\min_{p, C} \left\{ \frac{1}{m} \sum_{i=1}^m \ell(p(C(x_i)), y_i) + \Omega(C, p; x_1, \dots, x_m, x)\right\}.
\end{align*}

Intuitively, we want to find a raw data to symbols mapping $\hat{C}$ and program $\hat{p}$ which perform well on a training set (if one is available) and minimize the loss $\Omega$ over the given test samples. Notably, the program and raw data to symbol mapping do not need to be general, they should be \textit{specialized} for the test samples. In practice, the raw data to symbol mapping is often a form of semantic parsing, which can be done through zero-shot prompting of foundation models, so the raw data to symbol mapping is not modified on an instance-level.
 
\end{document}